\documentclass[manuscript,screen,review=false]{acmart}

\AtBeginDocument{%
  \providecommand\BibTeX{{%
    \normalfont B\kern-0.5em{\scshape i\kern-0.25em b}\kern-0.8em\TeX}}}

\setcopyright{acmcopyright}
\copyrightyear{2018}
\acmYear{2018}
\acmDOI{XXXXXXX.XXXXXXX}

\usepackage{amsmath}
\usepackage{amsfonts}

\usepackage{graphicx}
\usepackage{subcaption}

\usepackage{pgfplots}
\pgfplotsset{compat=newest}

\usepackage{longtable}
\usepackage{multirow}
\usepackage{changepage}

\usepackage{csquotes}

\usepackage{listings}
\usepackage{color}
\definecolor{backcolor}{RGB}{250, 250, 250}
\definecolor{cmtcolor}{RGB}{140, 250, 140}
\definecolor{kwcolor}{RGB}{60, 150, 240}
\definecolor{numcolor}{RGB}{120, 120, 120}
\definecolor{strcolor}{RGB}{240, 210, 60}
\usepackage{cite}
\usepackage{comment}
\usepackage{xcolor}
\usepackage{soul}

\begin{document}

%\title[Bio-Inspired Deep Learning]{Bio-Inspired Approaches for Deep Learning: A Survey on Synaptic Plasticity and Unsupervised Pattern Discovery}
%\title[Bio-Inspired Deep Learning]{Bio-Inspired Models of Synaptic Plasticity and Unsupervised Deep Learning: A Survey}
%\title[Bio-Inspired Deep Learning]{Synaptic Plasticity and Unsupervised Models for Bio-Inspired Deep Learning: A Survey}
\title[Bio-Inspired Deep Learning]{Synaptic Plasticity Models and Bio-Inspired Unsupervised Deep Learning: A Survey}

\author{Gabriele Lagani}
\authornote{Corresp.}
\email{gabriele.lagani@isti.cnr.it}
\author{Fabrizio Falchi}
\email{fabrizio.falchi@isti.cnr.it}
\author{Claudio Gennaro}
\email{claudio.gennaro@isti.cnr.it}
\author{Giuseppe Amato}
\email{giuseppe.amato@isti.cnr.it}
\affiliation{%
  \institution{ISTI-CNR}
  \city{Pisa}
  \country{Italy}
  \postcode{56124}
}

\renewcommand{\shortauthors}{Lagani, et al.}

\begin{abstract}
  Recently emerged technologies based on Deep Learning (DL) achieved outstanding results on a variety of tasks in the field of Artificial Intelligence (AI). However, these encounter several challenges related to robustness to adversarial inputs, ecological impact, and the necessity of huge amounts of training data. In response, researchers are focusing more and more interest on biologically grounded mechanisms, which are appealing due to the impressive capabilities exhibited by biological brains. 
  This survey explores a range of these biologically inspired models of synaptic plasticity, their application in DL scenarios, and the connections with models of plasticity in Spiking Neural Networks (SNNs). Overall, Bio-Inspired Deep Learning (BIDL) represents an exciting research direction, aiming at advancing not only our current technologies but also our understanding of intelligence.
\end{abstract}

\begin{CCSXML}
<ccs2012>
   <concept>
       <concept_id>10010147.10010257.10010293.10011809</concept_id>
       <concept_desc>Computing methodologies~Bio-inspired approaches</concept_desc>
       <concept_significance>300</concept_significance>
       </concept>
   <concept>
       <concept_id>10010147.10010257.10010293.10011809</concept_id>
       <concept_desc>Computing methodologies~Bio-inspired approaches</concept_desc>
       <concept_significance>500</concept_significance>
       </concept>
 </ccs2012>
\end{CCSXML}

\ccsdesc[300]{Computing methodologies~Bio-inspired approaches}
\ccsdesc[500]{Computing methodologies~Bio-inspired approaches}

\keywords{Bio-Inspired, Hebbian, Deep Learning, Neural Networks, Spiking}

%\received{}
%\received[revised]{}
%\received[accepted]{}

\maketitle

\section{Introduction}

In the past decade, Deep Learning (DL) technologies have attained performance levels equivalent to, or even surpassing, human capabilities across a multitude of Artificial Intelligence (AI) applications, such as computer vision \citep{he2016}, reinforcement learning \citep{silver2016}, or language processing \citep{devlin2019}. 
Although Deep Neural Network (DNN) models were originally inspired by biological mechanisms, current technologies have observed a significant departure from their biological counterparts. For example, the biological plausibility of the error backpropagation algorithm (\textit{backprop}) -- the workhorse of DL -- is questioned by neuroscientists \citep{oreilly, marblestone2016, hassabis2017, lake2017, richards2019}.

In this survey, we highlight the connections of biological plausibility with a number of other challenges that current DL solutions still need to overcome, such as the lack of robustness of traditional DNN architectures to adversarially perturbed inputs \citep{goodfellow2014b}, the necessity of huge amounts of labeled data \citep{roh2019}, or the ecological impact of neural network training \citep{badar2021}. For example, in \citep{badar2021} it is shown how more complex and large-scale models allow to improve benchmark results but at a higher energy cost, which is typically not taken into account when comparing against other models. 
On the other hand, biological intelligence can exhibit proficient and robust behavior in a variety of tasks \citep{mainen1995, gerstner1996}, while generalizing from little experience \citep{lake2020}, with an exceptionally low energy expenditure \citep{javed2010}.
Therefore, it appears that drawing inspiration from biology could once again provide valuable insights toward addressing the challenges presented above. Indeed, in recent years, significant research efforts have been devoted to the development of bio-inspired solutions for DL.

In order to achieve a better understanding of the principles and mechanisms behind biological intelligence, scientific investigation moves from two different perspectives. On one hand, neuroscientists uncover the low-level working principles of biological intelligent systems and try to relate them to high-level intelligent behavior in a bottom-up fashion. On the other hand, computer scientists start from high-level abstractions to model AI problems and then work out the structures and architectural details needed to solve such problems. Unfortunately, finding the connections between the high-level and the low-level aspects is often difficult.

\begin{figure*}
    \centering
    \includegraphics[width=0.3\textwidth]{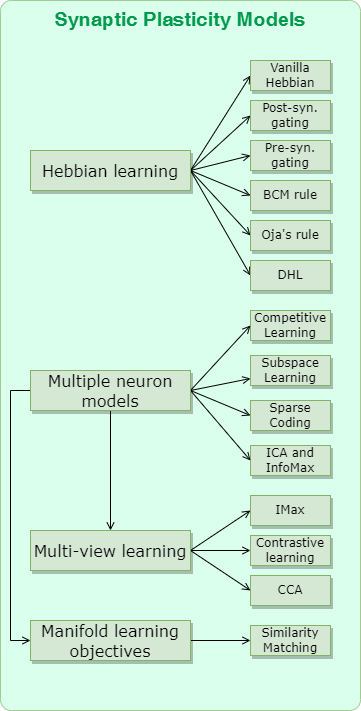}
    \caption{A schematic view of the topics of Bio-Inspired Deep Learning (BIDL) addressed in this work. 
    We provide a comprehensive discussion on biologically grounded synaptic plasticity models, starting from Hebbian learning models for a single neuron to more complex models for populations of multiple neurons, such as competitive learning, subspace learning, etc. These approaches are related to pattern discovery mechanisms such as clustering, Principal Component Analysis (PCA), manifold learning, etc., thus providing interesting connections between neuroscientific models and computer science/engineering aspects of AI. One of the goals of this survey is to highlight the relationships between these two fields, showing how complex intelligent behavior can emerge from biologically-inspired principles.
    }
    \label{fig:bidl}
\end{figure*}

The goal of this survey is to provide a comprehensive review of Bio-Inspired Deep Learning (BIDL), highlighting the connections between neuroscience and computer science viewpoints. We aim to provide a complete picture of the variety of perspectives, methods, and solutions that come together in this field, ranging from synaptic plasticity models to biologically realistic Spiking Neural Network (SNN) models. 
Fig. \ref{fig:bidl} provides a schematic summary of the various sub-topics embraced in this document.

This survey is intended both as a first approach for readers that are new to this field as well as a compact reference for a more experienced audience. The contents of this document should be easily accessible to computer scientists, as no prerequisite neuroscientific knowledge is expected, but they could also represent an interesting read for neuroscientists that are curious about the engineering aspects behind AI.

This document is structured as follows:
\begin{itemize}
    \item Section \ref{sec:related} discusses related surveys in the BIDL field.
    \item Section \ref{sec:plasticity_hebbian} describes biological synaptic plasticity models based on the Hebbian principle for a single neuron.
    \item Section \ref{sec:plasticity_multineuron} illustrates the extensions of plasticity mechanisms to populations of multiple neurons. It will be interesting to observe how certain biological aspects turn out to be related to unsupervised pattern discovery methods, thus showing a surprising connection between computer science and neuroscience.
    \item Section \ref{sec:plasticity_dl} presents some experimental results from the literature regarding the application of synaptic plasticity models to DL contexts and the integration of traditional backprop-based learning and bio-inspired synaptic plasticity.
    \item Section \ref{sec:spiking} introduces biologically detailed models of neural computation based on SNNs, and highlights their technological potential for energy-efficient neuromorphic computing.
    \item Finally, we present our conclusions in Section \ref{sec:concl}, outlining open challenges and possible future research directions.
\end{itemize}

Moreover, in our companion paper \citep{survey_snn}, we discuss biological models of spike coding and computation in greater detail, and we highlight the challenges of training such models with traditional backprop-based optimization. Therefore, we discuss recently proposed training algorithms, which pose themselves as alternatives to backprop, both for spiking and traditional architectures. These novel perspectives are promising to enhance the learning capabilities of current DL systems through biological insights and inspiration.

\section{Related Surveys} \label{sec:related}

Several aspects of BIDL have been reviewed in past contributions, each focused on a specific area or collection of methods. Some of these contributions are rooted in the neuroscientific viewpoint, while others explore more practical perspectives on DL system engineering. Our contribution aims at achieving a comprehensive presentation of the various domains and their interplay involved in the BIDL field, showing the connections between the neuroscientific aspects and the engineering abstractions.

An interesting treatment on the relationships between neuroscience and AI can be found in \citep{marblestone2016}, where the authors outline successful contributions in various aspects of DL, and propose directions of investigation for the neuroscientific field based on the insights provided by the high-level computational abstractions engineered by computer scientists. Similarly, the authors also suggest possible directions of inspiration that might come from neuroscience, towards the development of novel DL solutions. A similar discussion is presented in \citep{hassabis2017}, where the authors review historical interactions between computer science and neuroscience, showing how these interactions lead to novel results in these fields, and highlight possible shared themes for future development. Conversely, in \citep{richards2019} it is outlined how architectures, objective functions, and learning rules developed for artificial learning systems can inspire further developments in system neuroscience. Focusing on biological and psychological inspiration, another work \citep{lake2017} suggests specific areas of exploration towards building more human-like DL systems, ranging from causal reasoning to compositional learning and program induction, as well as learning-to-learn approaches.

Concerning the biologically grounded modeling of neural systems, in their book, Gerstner and Kistler \citep{gerstner} provides a comprehensive presentation of SNN models, as well as Hebbian plasticity and Spike Time Dependent Plasticity (STDP) models. A variety of Hebbian plasticity models are also reviewed in \citep{gorchetchnikov2011, vasilkoski2011}, while recent SNN developments and applications are surveyed in \citep{pfeiffer2018, tavanaei2019a, nunes2022}.

Compared to previous surveys, we provide significant contributions in the following directions:
\begin{itemize}
    \item Compared to works more focused on high-level perspectives about the interplay between neuroscience and computer science as a source of biological inspiration \citep{marblestone2016, hassabis2017, richards2019, lake2017}, our work delves deeper into specific aspects where the connections between neuroscientific models and emergent computational properties arise;
    \item We provide a comprehensive description of bio-inspired synaptic plasticity models, showing the connections with learning principles that lead to autonomous pattern discovery as a resulting behavior, while other contributions only focus on the biological models of synaptic plasticity \citep{gerstner, gorchetchnikov2011, vasilkoski2011};
    \item We provide parallel perspectives on bio-inspired methods for traditional networks and those based on spiking models, compared to works dedicated only to low-level spike-based methods \citep{pfeiffer2018, tavanaei2019a, nunes2022}.
\end{itemize}

\section{Synaptic Plasticity Models and Hebbian Learning in a Single Neuron} \label{sec:plasticity_hebbian}

Throughout life, our brain is continually subject to modifications in order to incorporate knowledge from the environment and adapt to new tasks. This adaptation process is referred to as \textit{plasticity}. The most prominent form of plasticity occurs in synapses, in the form of strengthening of synaptic efficacy, a.k.a. Long Term Potentiation (LTP), or weakening, a.k.a. Long Term Depression (LTD) \citep{bear1996, gerstner}. Given the crucial role that synaptic plasticity plays in neural systems, we begin this review by discussing synaptic plasticity models, starting from the simplest formulation of Hebbian plasticity, and then moving to more complex models that also recently found applications in DNN training. One of the challenges of formulating biologically plausible models of synaptic plasticity is the necessity to define \textit{local} learning rules, i.e. using only information that is locally available at the neuron site for computing the synaptic updates.

\begin{figure}
    \centering
    \includegraphics[width=0.4\textwidth]{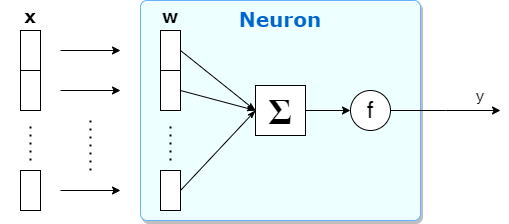}
    \caption{Representation of a neuron model which takes a vector $\mathbf{x}$ as input. Inputs are modulated by synaptic weights $\mathbf{w}$, and then summed together, before a nonlinearity $f(\cdot)$ is applied. The resulting output is $y = f(\sum_i w_i \, x_i)$ (where subscript $i$ indexes a specific vector entry).}
    \label{fig:neuron}
\end{figure}

Let us consider a neuron model whose synaptic weights are described by a vector $\mathbf{w}$. The neuron takes as input a vector $\mathbf{x}$, and produces an output $y(\mathbf{x}, \mathbf{w}) = f(\mathbf{x}^T \, \mathbf{w})$ (Fig. \ref{fig:neuron}), where $f$ is the activation function (optionally, a bias term can be implicitly modeled as a synapse connected to a constant input). In the following, we use boldface fonts to denote vectors, and normal fonts to denote scalars. In neuroscientific terms, input values $\mathbf{x}$ are also termed \textit{pre-synaptic} activations, while the output $y$ is termed \textit{post-synaptic} activation.

In order to model synaptic plasticity, neuroscientists propose the \textit{Hebbian} principle \citep{haykin, gerstner}: \textit{"fire together, wire together".} According to this principle, the synaptic coupling between two neurons is reinforced when the two neurons are simultaneously active \citep{haykin, gerstner}. Mathematically, this learning rule, in its simplest "vanilla" formulation, can be expressed as:
\begin{equation} \label{eq:hebbian_vanilla}
     \Delta w_i = \eta \, y \, x_i
\end{equation}
where $\eta$ is the learning rate, and the subscript $i$ refers to the i-th input/synapse.
The effect of this learning rule is essentially to consolidate correlated activations between neural inputs and outputs, by reinforcing the synaptic couplings, so that, if a similar input will be observed again in the future, a similar response will likely be elicited from the neuron.
We can already outline a first connection between the neuroscientific Hebbian learning theory and data science, specifically Principal Component Analysis (PCA): if multiple inputs are presented to a neuron, assuming that the activation function is linear and that the inputs have zero means, it can be shown \citep{gerstner, oja1982} that Eq. \ref{eq:hebbian_vanilla} induces the weight vector to align towards the principal component of the data distribution.

\begin{figure}
    \centering
    \begin{subfigure}[t]{0.3\textwidth}
        \includegraphics[width=\textwidth]{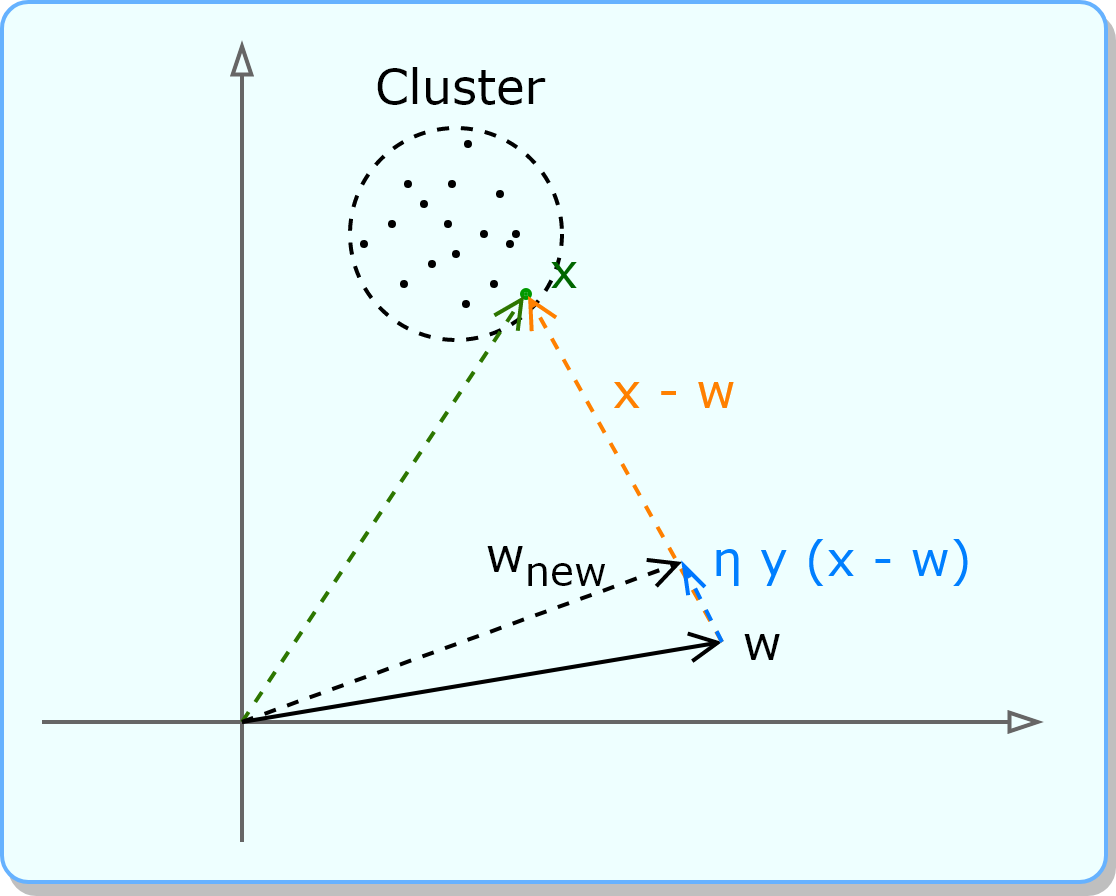}
        \caption{Weight vector subject to an update. Points are inputs (organized in a cluster), and the green point is the input currently being processed. The blue arrow represents the direction of the update that will affect the weight vector $w$, while the red arrow is the actual update.}
        \label{fig:x-w_update_visualized_a}
    \end{subfigure}
    ~
    \begin{subfigure}[t]{0.3\textwidth}
        \includegraphics[width=\textwidth]{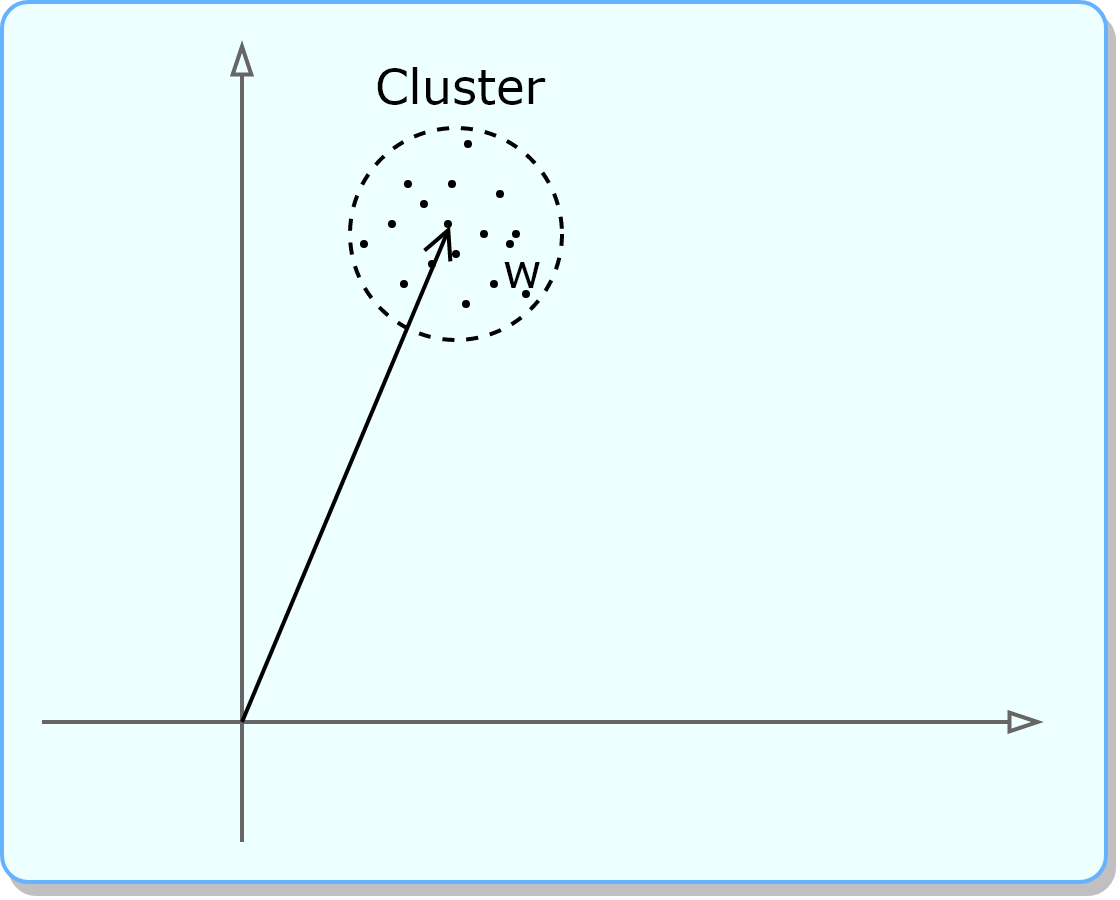}
        \caption{Final position of the weight vector after training.}
        \label{fig:x-w_update_visualized_b}
    \end{subfigure}
    \caption{Effect of Hebbian updates on a weight vector.}
    \label{fig:x-w_update_visualized}
\end{figure}

The problem with Eq. \ref{eq:hebbian_vanilla} is that there are no mechanisms to prevent the weights from growing unbounded, thus leading to possible instability. This issue can be counteracted by adding a weight decay term $\gamma(\mathbf{w}, \mathbf{x})$ to the learning rule:
\begin{equation} \label{eq:hebbian_wd}
     \Delta w_i = \eta \, y \, x_i - \gamma(\mathbf{w}, \mathbf{x})
\end{equation}
In particular, with an appropriate choice for this term, i.e. $\gamma(\mathbf{w}, \mathbf{x}) = \eta \, y(\mathbf{w}, \mathbf{x}) \, w_i$, we obtain a learning rule that has been widely applied to the context of \textit{competitive} learning (which will be discussed more in detail later) \citep{grossberg1976a, rumelhart1985, kohonen1982}:
\begin{equation} \label{eq:hebbian_centroid}
     \Delta w_i = \eta \, y \, x_i - \eta \, y \, w_i = \eta \, y \, (x_i - w_i)
\end{equation}
This rule has a simple intuitive interpretation, depicted in Fig. \ref{fig:x-w_update_visualized}: when a collection of inputs is presented to the neuron, the weight vector evolves towards the centroid of the cluster formed by the inputs. In essence, the neuron is storing, in its synapses, a prototypical representation of the patterns observed during training. When a similar pattern is presented again, the neuron will produce a stronger response, thus becoming a sort of pattern-matching unit.

The learning rule above is a special case of \textit{post-synaptic gating} \citep{gerstner}:
\begin{equation} \label{eq:hebbian_postgate}
     \Delta w_i = \eta \, y \, (x_i - \theta)
\end{equation}
where the update step $x_i - \theta$ is "gated" by the post-synaptic activation $y$. The parameter $\theta$ is a \textit{threshold} parameter that determines the behavior of the updates: if the pre-synaptic stimulus is stronger than the threshold, LTP will be induced, otherwise we have LTD. Conversely, it is possible to define a \textit{pre-synaptic gating} rule \citep{gerstner}, by inverting the role of $x_i$ and $y$:
\begin{equation} \label{eq:hebbian_pregate}
     \Delta w_i = \eta \, x_i \, (y - \theta)
\end{equation}
%where $\odot$ denotes element-wise product, and scalar-vector sum/subtraction should also be interpreted element-wise. 
The threshold parameter can be fixed, or it can depend on the weights, on the current pre/post-synaptic activations, or even on the history of past activations.

A mixed approach is taken in the \textit{covariance rule} \citep{sejnowski1989}, where thresholds are imposed both on the pre-synaptic and post-synaptic signals:
\begin{equation} \label{eq:hebbian_cov}
     \Delta w_i = \eta \, (y - \theta_y) \, ( x_i - \theta_x)
\end{equation}
Specifically, $\theta_x$ and $\theta_y$ are running averages of pre- and post-synaptic activities over time, which are biologically supported by the concept of \textit{synaptic traces} \citep{izhikevich2007, yagishita2014, shindou2019, gerstner2018}. The covariance rule adapts vanilla Hebbian learning to the case of data with non-zero mean, tracking statistics online in the same spirit as \textit{batch normalization} \citep{ioffe2015}.

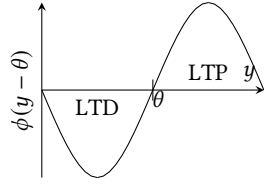
\begin{figure}
\centering
\begin{tikzpicture}
\begin{axis}[
    width=0.3\textwidth,
    axis x line=middle,
    axis y line=left,
    xlabel={$y$},
    ylabel={$\phi (y - \theta)$},
    ticks=none,
    samples=10,
]
    \addplot[][domain=0:0.5]{+4*x^3+2.92e-60*x^2+-3*x^1+0*x^0};
	\addplot[][domain=0.5:1]{+-4*x^3+12*x^2+-9*x^1+1*x^0};
	\addplot[][domain=1:1.5]{+-4*x^3+12*x^2+-9*x^1+1*x^0};
	\addplot[][domain=1.5:2]{+4*x^3+-24*x^2+45*x^1+-26*x^0};
	\addplot[mark=|,mark size=4,mark options={solid}][] coordinates{(1, 0)};
	\node[] at (axis cs: 1.05, -0.1) {$\theta$};
	\node[] at (axis cs: 0.5, -0.2) {LTD};
	\node[] at (axis cs: 1.5, 0.2) {LTP};
\end{axis}
\end{tikzpicture}
\caption{Non-linearity in BCM rule.}
\label{graph:bcm_phi} 
\end{figure}

The \textit{Bienenstock-Cooper-Munro} (BCM) rule \citep{bienenstock1982} introduces a nonlinearity $\phi$ in the learning process: 
\begin{equation} \label{eq:hebbian_bcm}
     \Delta w_i = \eta \, x_i \, \phi(y - \theta)
\end{equation}
Fig. \ref{graph:bcm_phi} shows a typical shape of the nonlinearity.
A threshold is still applied to the post-synaptic activity, so that LTP occurs when the neuron activity is above the threshold, or LTD takes place otherwise. However, when the activity becomes too large, too small, or in proximity to the threshold, plasticity is inhibited. The idea is that, when neural activity is too large (or too small), LTP leads to a further increase in the activity (conversely, LTD leads to a further decrease), but the nonlinearity provides a stabilizing effect.

Another approach for setting bounds on the synaptic weights could be to adopt \textit{soft thresholds} on the weights, i.e. explicit terms in the update equations that automatically limit the weight updates when a given threshold is approached \citep{gerstner}. 
\begin{equation} \label{eq:hebbian_soft_thr}
     \Delta w_i = \eta \, \Delta w_i^{(base)} \, (w_i - \theta_{LB}) \, (\theta_{UB} - w_i)
\end{equation}
where $\theta_{LB}$ and $\theta_{UB}$ act as soft lower and upper thresholds for the weights, with $\theta_{LB} < \theta_{UB}$, and $\Delta w_i^{(base)}$ is a given weight update before soft-thresholding.
A similar concept also arises in \textit{bi-stable synapse} models \citep{fusi2000, gerstner}. In these models, synaptic weights are allowed to settle down only in one of two possible stable states (for example, 0 and 1). The learning rule is divided into two parts:
\begin{equation} \label{eq:hebbian_bistable}
     \Delta w_i = H + R
\end{equation}
where $H$ is a generic Hebbian term driving plasticity, while $R$ is a \textit{refresh} term for synaptic stabilization:
\begin{equation} \label{eq:bistable_refresh}
    R = \gamma \, w_i \, (1 - w_i) \, (w_i - \theta)
\end{equation}
where $\gamma$ is a constant. This term drives weight values towards 1, when the current weight value is above a threshold $\theta$, or to 0 otherwise. 
%The actual weight value used for computations is 1, when a given weight is above the threshold, or 0 when it is below. 
%The purpose of this rule is to tackle the \textit{stability-plasticity dilemma}: if new patterns are presented to the neuron, plasticity might cause the neuron to forget what was learned in the past, causing what is known as \textit{catastrophic forgetting}.
Thanks to this rule, transitions of the weight value from 0 to 1 can occur only when the driving Hebbian term provides a stimulation strong enough to bring the weight value beyond the threshold, and vice-versa for transitions from 1 to 0. 

A different approach to the weight instability problem has been proposed in \citep{oja1982}: the idea is to renormalize the weight vector after each update, so that its length is always kept constant, although the direction changes over time, aligning towards the data principal component (under the same assumptions as in the vanilla case). It can be shown that, under small learning rates, the learning rule with the renormalization step can be approximated by a first-order Taylor expansion around $\eta = 0$, leading to the weight update named \textit{Oja's rule}:
\begin{equation} \label{eq:hebbian_oja}
     \Delta w_i = \eta \, y \, ( x_i - y \, w_i)
\end{equation}
Note that this update also falls in the category of Hebbian updates with weight decay, which in this case is $\gamma(\mathbf{w}, \mathbf{x}) = \eta \, y(\mathbf{w}, \mathbf{x})^2 \, w_i$.

It is worth noting that the aforementioned learning rules can be interpreted as instances of a more general \textit{local} synaptic update equation which, for a generic synaptic connection $i$, can be expressed as \citep{gerstner}:
\begin{equation} \label{eq:hebbian_general}
     \Delta w_i = a_0 + a_1 x_i + a_2 y + a_3 x_i y + a_4 x_i^2 + a_5 y^2 + ...
\end{equation}
where the coefficients $a_i$ may depend on the weights.

\textit{Differential Hebbian Learning} (DHL) \citep{kosko1986} represents a departure from traditional Hebbian models. Instead of considering simply the pre- and post-synaptic activities at a given instant, this model proposes to consider also the rate of change of these activities over time:
\begin{equation} \label{eq:hebbian_differential}
    \Delta w_i = \eta \, \frac{d y}{d t} \, x_i
\end{equation}
This learning rule has interesting properties related to data decorrelation \citep{choi1998}, temporal difference learning \citep{kolodziejski2009a, kolodziejski2009b}, and encoding of error signals in STDP neurons \citep{roberts1999} providing a biologically grounded mechanism for Contrastive Hebbian Learning (CHL) \citep{movellan1991} in networks of spiking neurons \citep{bengio2015b} (we will come back to these topics in the following Sections).

The learning rules presented so far involve only a single neuron. In the next subsection, we will consider more complex learning scenarios involving multiple neurons.

\section{Plasticity Models for Unsupervised Pattern Discovery with Multiple Neurons}  \label{sec:plasticity_multineuron}

So far, we have considered plasticity models for single neurons. When we are dealing with a population with multiple neurons, naively applying a synaptic update rule such as those from the previous subsection is not guaranteed to be effective for a learning task. In fact, if multiple neurons follow the same learning dynamics, it is easy for them to converge to similar configurations. It is instead desirable to achieve some form of \textit{decorrelation} of neural activity, i.e. making sure that different neurons learn to encode different pieces of information for given inputs \citep{foldiak1990, olshausen1996a}, in order to maximize the representation power. This subsection explores the strategies to achieve such decorrelation in neural populations with local synaptic plasticity.

\subsection{Background on Neural Cells}

Let us introduce some biological background on the various types of neural cells and their functions, from which it will be possible to draw relationships between the computational learning mechanisms that we are going to discuss in the following, and the biological substrate that supports such mechanisms.

Neural cells can be classified into two main groups: pyramidal cells and non-pyramidal cells \citep{white}.

Pyramidal cells represent the fundamental computing unit of biological neural networks. They typically have a pyramid-like shape, and they are endowed with two types of \textit{dendrites} (i.e. input connections): \textit{apical} dendrites, which extend from the tip of the pyramid, and several \textit{basal} dendrites, originating from the base. Apical dendrites extend through cortical layers, while basal dendrites extend mainly toward neighboring cells in the same region.
The \textit{axon} of pyramidal cells (i.e. the output connection) originates at the base of the pyramid and it immediately branches into a \textit{projection} axon and several \textit{axon collaterals}. The projection axon extends towards deeper layers. Axon collaterals can be \textit{local}, i.e. extending for a short distance towards neighboring neurons, or they can extend for longer distances either in the same layer or towards other layers.

Concerning the non-pyramidal cells, they can be further divided into a variety of categories \citep{stefanis2020}, but the common features are a central body with a smaller size compared to pyramidal cells, a number of dendrites originating from it, and an axon which tends to branch into multiple ramifications. Dendrites and connections are mainly local, hence these cells tend to connect to neighboring neurons, such as pyramidal cells, thus transmitting information about the activity in the neighborhood. By virtue of this role, these cells are often labeled as \textit{interneurons}.
Axons of non-pyramidal cells tend to form mainly inhibitory connections with other neural elements, which play an important role in inhibitory interaction and \textit{shunting inhibition} \citep{kubota2016} between neurons.

In the following, we will highlight the role played by artificial neurons, which correspond to pyramidal cells, and the mechanisms of inhibitory \textit{lateral interaction}, mediated by non-pyramidal cells, which is essential to achieve the necessary decorrelation in neural activity, thus showing an interesting mapping between biological circuits and artificial learning systems.

\subsection{Competitive Learning}

\begin{figure}
    \centering
    \begin{subfigure}[t]{0.3\textwidth}
        \includegraphics[width=\textwidth]{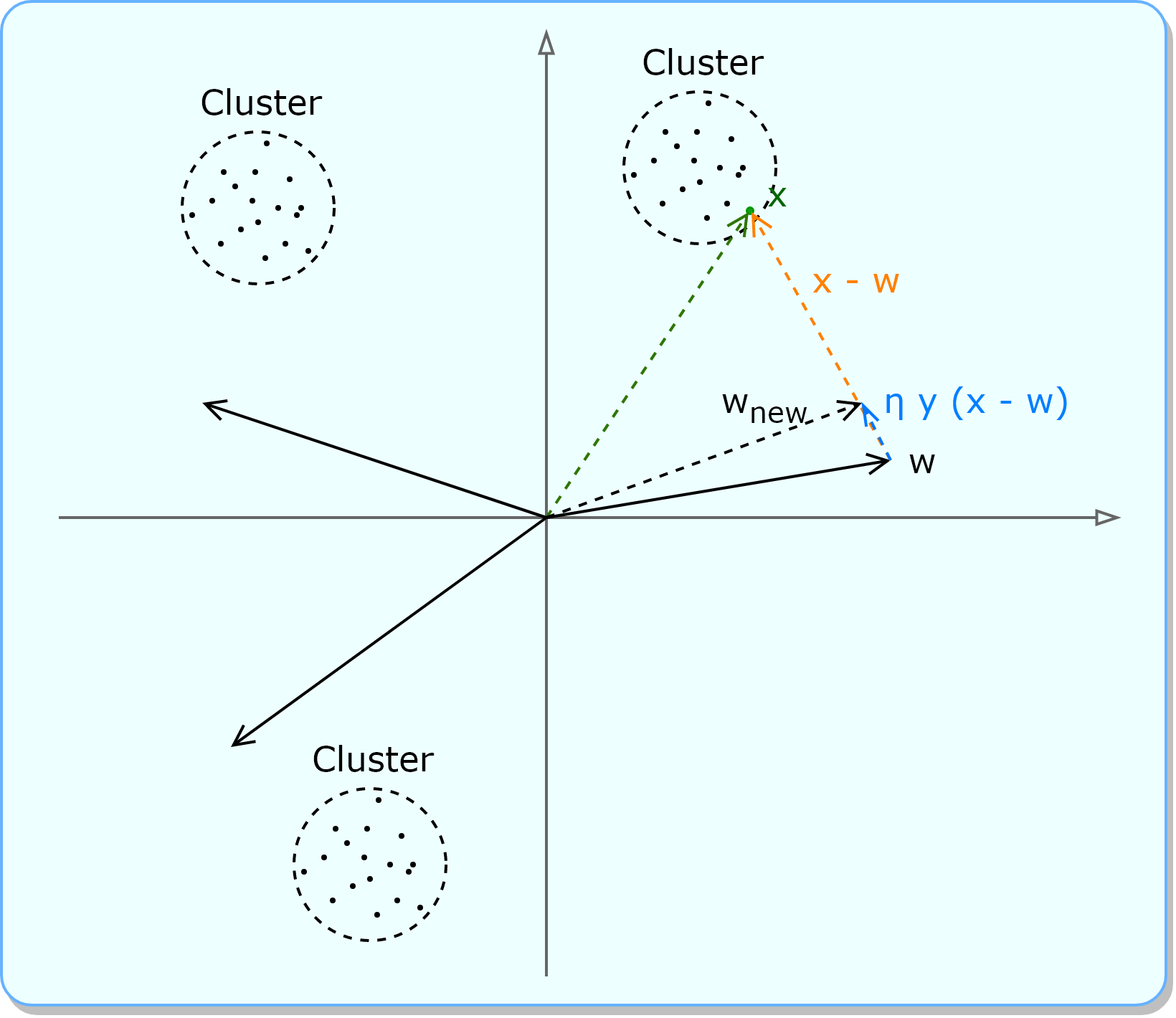}
        \caption{Update step}
        \label{fig:competitive_update_visualized_a}
    \end{subfigure}
    ~
    \begin{subfigure}[t]{0.3\textwidth}
        \includegraphics[width=\textwidth]{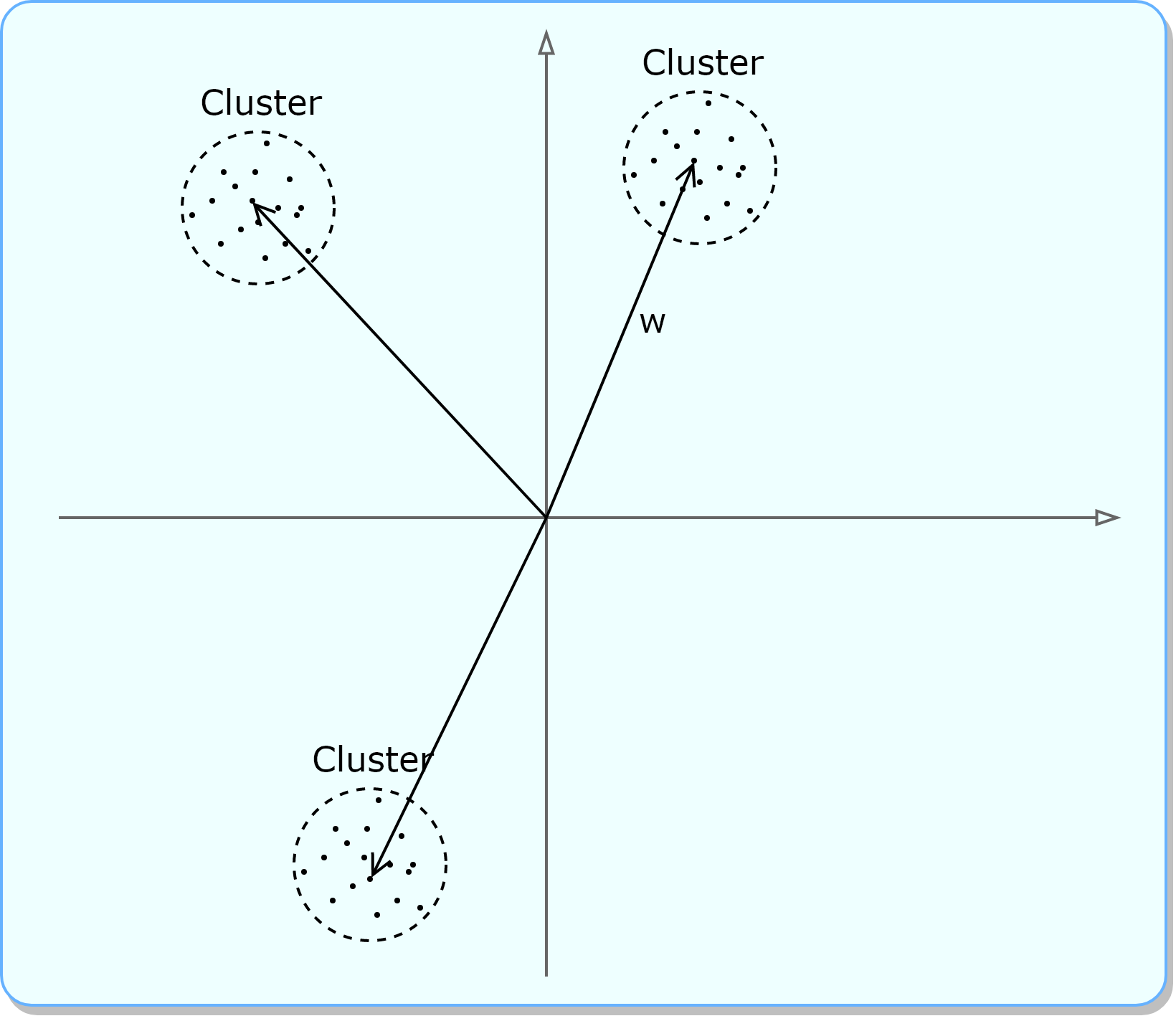}
        \caption{Final position after convergence}
        \label{fig:competitive_update_visualized_b}
    \end{subfigure}
    \caption{Hebbian updates with Winner-Takes-All competition.}
    \label{fig:competitive_update_visualized}
\end{figure}

When multiple neurons are involved in a complex network, \textit{competitive learning} can be adopted to force different neurons to learn different patterns. A possible strategy is \textit{Winner-Takes-All} (WTA) \citep{grossberg1976a, rumelhart1985}: when an input is presented to a WTA layer, the neuron whose weight vector is closest to the current input (e.g. in terms of angular or euclidean distance) is elected as \textit{winner}. Only the winner is allowed to perform a weight update, according to Eq. \ref{eq:hebbian_centroid}, thus moving its weight vector closer to the current input (Fig. \ref{fig:competitive_update_visualized}). If a similar input will be presented again in the future, the same neuron will be more likely to win again. Competitive interaction between neurons is biologically motivated by the lateral inhibition mechanisms \citep{gabbot1986} that we have previously highlighted. This strategy allows a group of neurons to align their weight vectors towards the centroids of distinct clusters formed by the data points (Fig. \ref{fig:competitive_update_visualized}), which shows another connection between a neuroscience-inspired learning theory, and a data analytic operation, namely clustering.

The following equation gives a mathematical description of Hebbian WTA learning:
\begin{equation} \label{eq:hwta}
    \Delta w_{i, j} = \eta \, r_j \, (x_{i} - w_{i, j})
\end{equation}
Here, subscripts $i$ and $j$ refer to the i-th input/synapse and j-th neuron in the layer, respectively, and $r_j$ is the neuron activation after a competitive nonlinearity: it is $1$ for the winning neuron and $0$ otherwise. 
A variant of WTA is k-WTA \citep{majani1989}, where top-k neurons closest to the input are the winners, which means that $r_j = 1$ for these neurons and $0$ for all the others.

In contrast with the sharp competition provided by WTA, soft forms of competition are also possible. In these cases, instead of having sharp winning neurons, corresponding to $r_j$ being $0$ or $1$, we can also attribute intermediate values to $r_j$ for each neuron. For example, soft-WTA \citep{nowlan1990} allows all the neurons to receive a score based on their activations so that neurons with higher activation will receive a higher score. In the original work, the score was computed simply as an $L_1$ normalization of the activations of the neurons. In \citep{lagani2021d, moraitis2021}, other soft-WTA variants were introduced where the score was computed as the $L_p$ normalization or as the softmax of neural activations, i.e.:
\begin{equation}
    r_j = \frac{y_j^p}{\sum_k y_k^p}
\end{equation}
\begin{equation}
    r_j = \frac{e^{y_j/T}}{\sum_k e^{y_k/T}}
\end{equation}
respectively.
In the latter equation, T is the \textit{temperature} parameter of the softmax \citep{gao2017}, which serves to cope with the variance of the activations.
Note that $r$ can be viewed as a step modulation coefficient: the higher the neuron activity, the larger the update step will be.

\begin{figure}
    \centering
    \begin{subfigure}[t]{0.15\textwidth}
        \includegraphics[width=\textwidth]{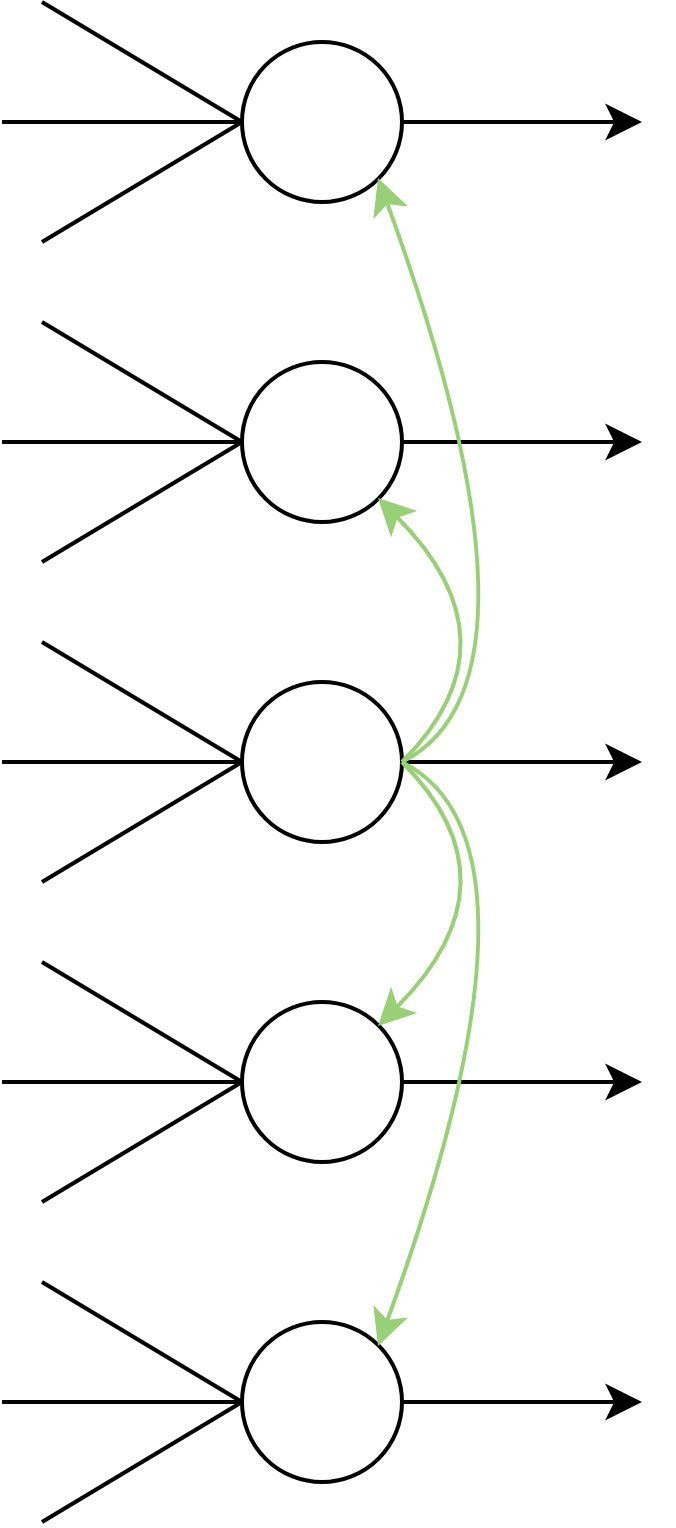}
        \caption{1-dimensional lattice.}
        \label{fig:som_topology_a}
    \end{subfigure}
    ~
    \begin{subfigure}[t]{0.3\textwidth}
        \includegraphics[width=\textwidth]{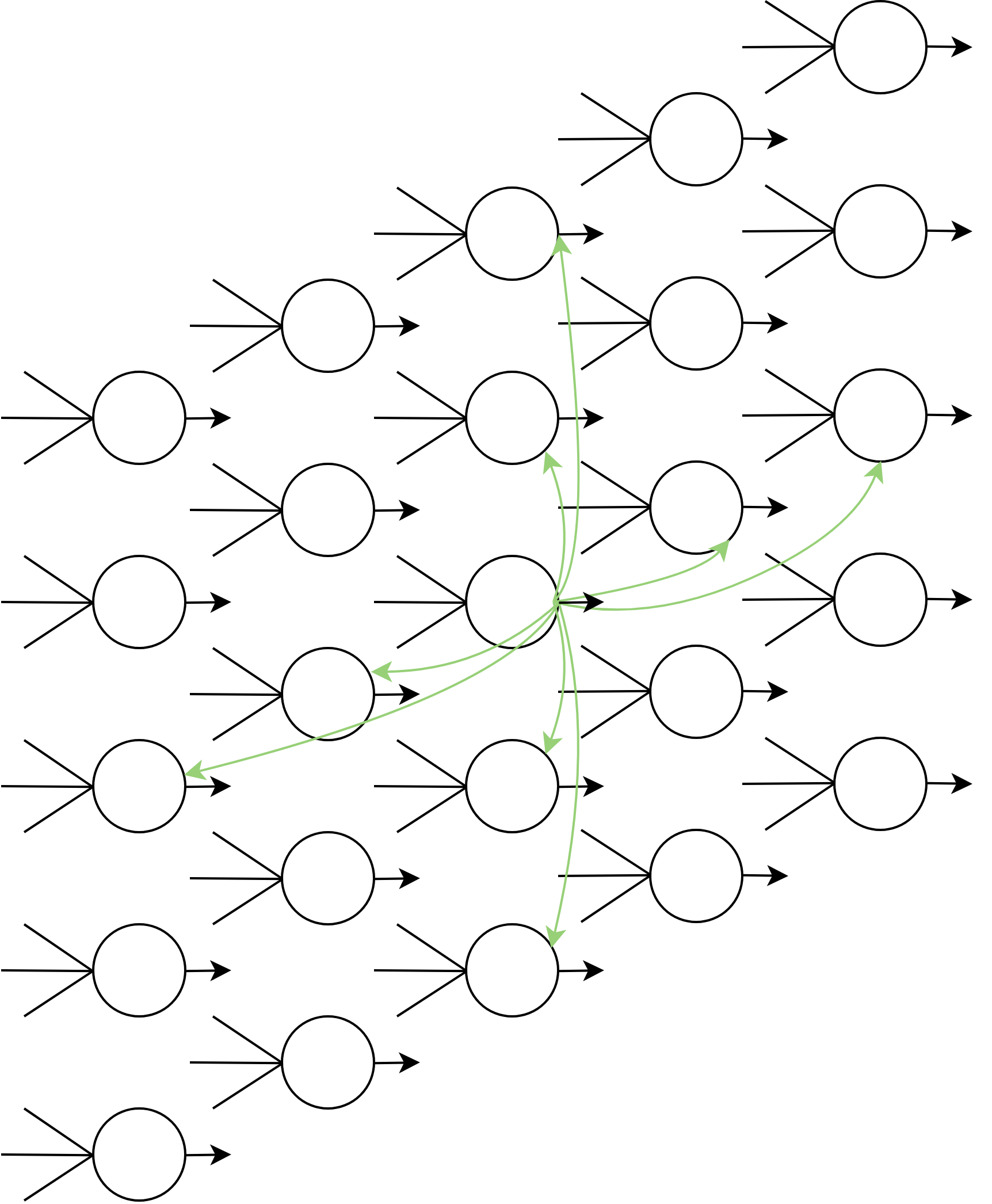}
        \caption{2-dimensional lattice.}
        \label{fig:som_topology_b}
    \end{subfigure}
    \caption[Self-Organizing Maps with neurons arranged in different topologies.]{Self-Organizing Maps with neurons arranged in different topologies. Some of the lateral feedback connections are highlighted in green.}
    \label{fig:som_topology}
\end{figure}

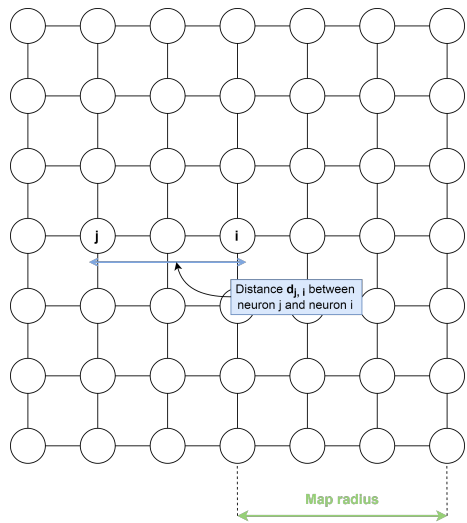
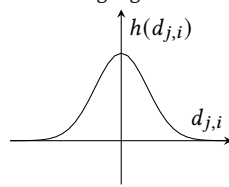
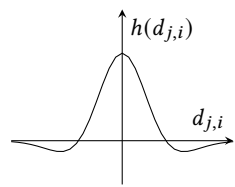
\begin{figure}
    \centering
    \begin{subfigure}[t]{0.4\textwidth}
        \includegraphics[width=\textwidth]{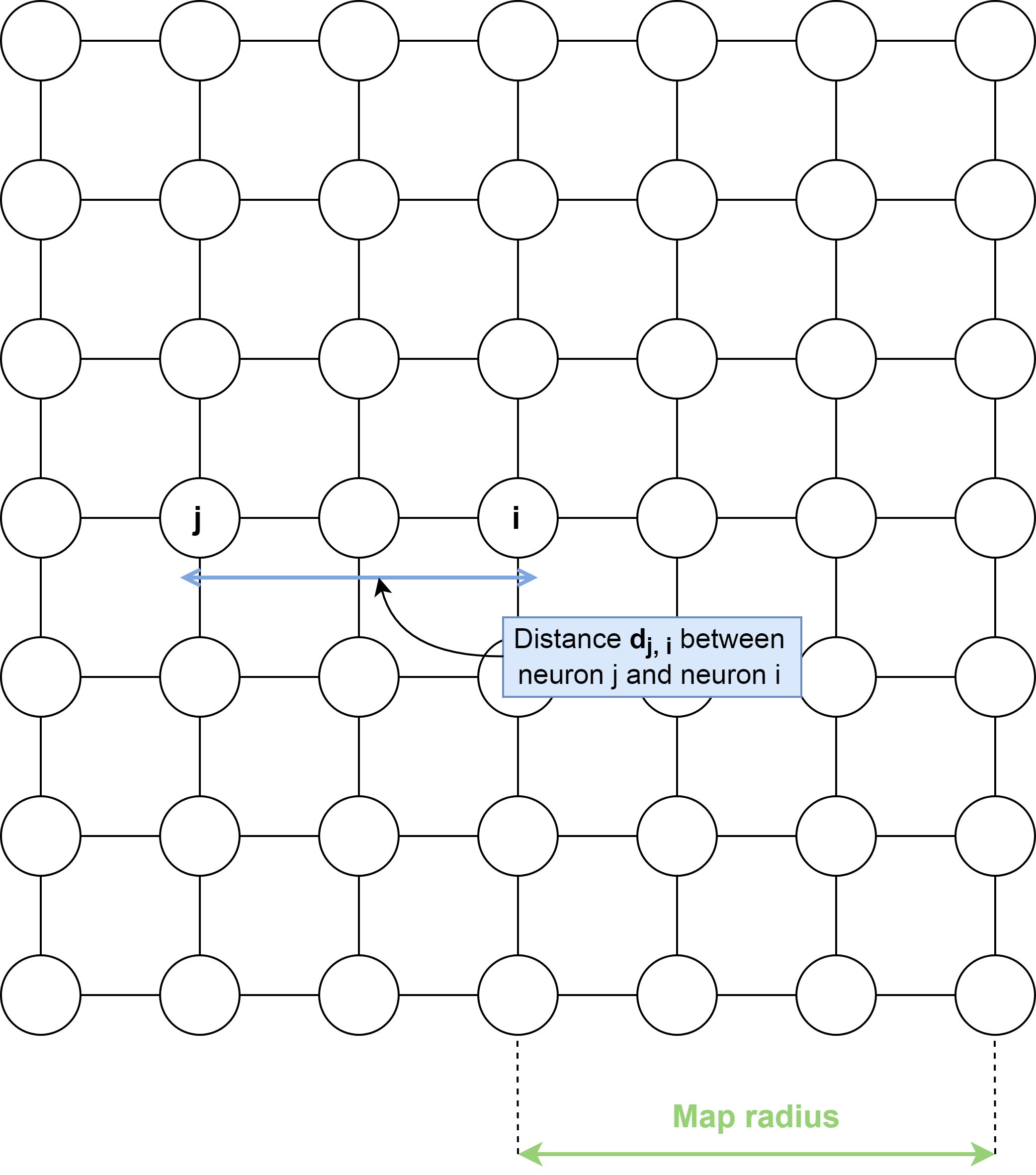}
        \caption{2-dimensional lattice with radius highlighted in green and distance $d_{j, i}$ between neuron $j$ and neuron $i$ highlighted in blue.}
        \label{fig:neighb_interact_a}
    \end{subfigure}
    \\
    \begin{subfigure}[t]{0.3\textwidth}
        \centering
        \begin{tikzpicture}
            \begin{axis}[
                axis x line=middle,
                axis y line=middle,
                xlabel = {$d_{j, i}$},
                ylabel = {$h(d_{j, i})$},
                ymax = 1.5,
                ymin = -0.5,
                width = \textwidth,
                ticks = none,
            ]
                \addplot[
                    domain=-3:3,
                    samples=35,
                ]{exp(-x^2)};
            \end{axis}
        \end{tikzpicture}
        \caption{Gaussian neighborhood function.}
        \label{fig:neighb_interact_b}
    \end{subfigure}
    ~
    \begin{subfigure}[t]{0.3\textwidth}
        \centering
        \begin{tikzpicture}
            \begin{axis}[
                axis x line=middle,
                axis y line=middle,
                xlabel = {$d_{j, i}$},
                ylabel = {$h(d_{j, i})$},
                ymax = 1.5,
                ymin = -0.5,
                width = \textwidth,
                ticks = none,
            ]
                \addplot[
                    domain=-3:3,
                    samples=35,
                ]{2*exp(-x^2) - exp(-(x^2)/2)};
            \end{axis}
        \end{tikzpicture}
        \caption{Mexican hat neighborhood function.}
        \label{fig:neighb_interact_c}
    \end{subfigure}
    \caption{Lateral interaction with some neighborhood function profiles.}
    \label{fig:neighb_interact}
\end{figure}

In \citep{kohonen1982}, the work on competitive learning was further extended with the introduction of \textit{Self-Organizing Maps} (SOMs). A SOM is a layer of neurons arranged in an n-dimensional lattice (typically 1-dimensional or 2-dimensional, the latter being more common). After the competitive phase, but before the weight update, training is extended with a new \textit{cooperative} phase, in which lateral interaction takes place in the form of a lateral feedback signal that is provided by the winning neuron to its neighbors in the lattice topology (fig. \ref{fig:som_topology}). The strength of this signal decreases with the distance from the winning neuron. Specifically, denoting with $i(x)$ the winning neuron on input $x$, the strength of the signal delivered to any neuron $j$, whose distance from $i(x)$ in the lattice topology is $d_{j, i(x)}$, is determined by the \textit{neighborhood function} $h(d_{j, i(x)})$. This function should be equal to $1$ when $d_{j, i(x)}$ is $0$ and should decrease with the distance. For instance, a possible choice for the neighborhood function can be a Gaussian function or a \textit{mexican hat} function centered in zero (Fig. \ref{fig:neighb_interact}) \citep{haykin, kohonen1993}. 
Other possible choices for the neighborhood function and further theoretical details about the SOMs are discussed in \citep{lo1991a, lo1993, erwin1991, erwin1992a, erwin1992b, cottrell2018}. 
The neighborhood function is characterized by a radius (the standard deviation in the Gaussian case) which is typically initialized to be equal to the radius of the lattice and then shrank over time. Once the cooperative phase is completed, the weight update takes place by applying eq. \ref{eq:hwta}, in which $r$ is set to $h(d_{j, i(x)})$. Note that the WTA approach can be seen as a particular case of SOM in which the neighborhood function has a radius of zero.

\subsection{Subspace Learning}

According to the definition given above, WTA enforces a kind of \textit{quantized} information encoding in neural network layers. Only one or a few neurons activate to encode the presence of a given pattern in the input. On the other hand, neural networks trained with backpropagation exhibit a \textit{distributed} representation \citep{agrawal2014}, where multiple neurons activate combinatorially to encode different properties of the input, resulting in an improved coding power. Similar distributed coding schemes have also been observed in biological neuron populations \citep{averbeck2006, wohrer2013}. The importance of distributed representations was also highlighted in \citep{foldiak1989, olshausen1996a}.

A more distributed coding scheme could be obtained by representing data as linear combinations of some orthonormal basis of feature vectors. Data projection over an orthogonal basis of weight vectors can be easily implemented as a neural mapping. In order to capture as much information as possible from the data, this basis should span the principal subspace, i.e. the subspace capturing most of the data variance, for which the data principal components form an orthogonal basis \citep{haykin}. 

A PCA-based neural network, PCANet, was proposed in \citep{chan2015}. Network filters were obtained by running PCA offline on the training dataset and using the extracted principal components as weights. Multiple processing layers were obtained by stacking PCA filters obtained from feature representations of the dataset extracted from previous layers, in a bottom-up fashion. The approach achieved 78\% accuracy on the CIFAR-10 \citep{cifar} dataset with a relatively shallow network. Unfortunately, Offline PCA computation is quite expensive, and it becomes prohibitive when applied to larger inputs or deeper networks. However, there exists an extension to Oja's rule that allows to perform PCA also in an online fashion, which is more appealing both in terms of efficiency and biological plausibility.

We have already observed how Oja's rule provides a stable Hebbian mechanism for extracting the first principal component. Further extensions of such mechanism for multiple neurons exist, which allow the extraction of successive directions spanning the principal subspace \citep{sanger1989, becker1996a}. 
In order to perform Hebbian PCA, a set of weight vectors has to be determined, for the various neurons, that minimize the \textit{representation error}, defined as:
\begin{equation} \label{eq:repr_err}
    \mathcal{L}_R(\mathbf{w_i})  = E[(\mathbf{x} - \sum_{j=1}^i y_j \, \mathbf{w_j})^2]
\end{equation}
where the subscript $i$ refers to the $i^{th}$ neuron in a given layer and $E[\cdot]$ is the mean value operator.
It can be pointed out that, in the case of linear neurons and zero-centered data, this reduces to the classical PCA objective of maximizing the output variance, with the weight vectors subject to orthonormality constraints \citep{sanger1989,becker1996a,karhunen1995}. 
From now on, let us assume that the input data are centered around zero. If this is not true, we just need to subtract the average from the inputs beforehand.

It can be shown that \textit{Sanger's rule} minimizes the objective in Eq. \ref{eq:repr_err} \citep{sanger1989}:
\begin{equation} 
    \Delta \mathbf{w_i} = \eta y_i (\mathbf{x} - \sum_{j=1}^i y_j \mathbf{w_j})
\end{equation}
The intuition behind this learning rule is the following: 1) for the first neuron, it simply corresponds to Oja's rule, thus extracting the first principal component; 2) for a generic successive neuron $i$, the learning rule subtracts from the input the partial representation $\sum_{j=1}^{i-1} y_j \mathbf{w_j}$ reconstructed from the previous neurons, thus canceling the subspace spanned by the first $i - 1$ principal components; 3) neuron $i$ then applies Oja's rule on the residual part of the input, which leads to the extraction of the $i$-th principal component.
In the case of nonlinear neurons, a solution to the problem can still be found \citep{karhunen1995}. Calling $f()$ the neuron activation function, under mild conditions that include the monotonic increase of $f$, the representation error
\begin{equation}
    \mathcal{L}_R(w_i)  = E[(\mathbf{x} - \sum_{j=1}^i f(y_j) \, \mathbf{w_j})^2]
\end{equation}
can be minimized with the following nonlinear version of the previous learning rule for nonlinear Hebbian PCA:
\begin{equation} \label{eq:hpca_lrn_rule}
    \Delta \mathbf{w_i} = \eta f(y_i) (\mathbf{x} - \sum_{j=1}^i f(y_j) \mathbf{w_j})
\end{equation}

%Nonlinear generalizations of the previous methods can also be achieved by means of nonlinear mappings into kernel spaces \citep{muller2001}, leading to algorithms such as Kernel PCA \citep{scholkopf1998, kim2003}. Kernel space mappings can be realized by leveraging Nystroem approximations \citep{williams2001b}. Clustering methods can be employed for this purpose \citep{zhang2010}, which are therefore well suited for biologically plausible implementations. Essentially, nonlinear extensions of linear Hebbian processing layers can be achieved by a pair of layers: a clustering layer to implement the kernel mapping through a Nystroem approximation, followed by a standard linear layer (Fig. \ref{fig:}).

Other variants of Hebbian PCA learning rules exist. The \textit{subspace learning} rule, also due to Oja \citep{oja1989, oja1992}, differs from Sanger's rule in that each neuron subtracts from the input the same reconstruction vector: $\sum_{j=1}^{N} y_j w_j$ (mind the summation index running from 1 to N, where N is the number of neurons in the layer). The resulting learning rule is:
\begin{equation}
    \Delta w_i = \eta y_i (x - \sum_{j=1}^{N} y_j w_j)
\end{equation}
With this variation in the learning scheme, the network is capable of extracting the \textit{principal subspace}, i.e. the same space spanned by the PCA directions, but expressed in terms of some other orthonormal basis (which is going to be a rotated version of the PCA basis).

Strictly speaking, Sanger's rule (as well as Oja's subspace rule) is not biologically plausible, because the weight update of a synapse is computed using information about weights on other synapses and outputs from other neurons (i.e. the terms in the sum). Nonetheless, network models can be designed that are functionally equivalent to Sanger's (or Oja's), but they are also consistent with biological plausibility requirements, using only local information in the updates. One of these was proposed in \citep{foldiak1989}, and similarly also in \citep{plumbley1993b}, which consisted of a linear single-layer network with both feedforward and lateral connections. Feedforward connections were trained with Oja's rule, while lateral connections were trained using an anti-Hebbian rule (i.e. with a minus sign in front), an update scheme known as \textit{Hebbian/anti-Hebbian} (HaH). Thanks to the lateral interaction, neurons were able to decorrelate their activity, projecting the data onto the principal subspace, while also normalizing the variance of neural activations. In data analysis, the operation performed by this network configuration corresponds to the \textit{whitening} transformation \citep{cifar}.

A similar approach was taken in \citep{rubner1989}, where purely Hebbian update was used for the feedforward connections and anti-Hebbian for the lateral ones, with explicit normalization performed after each update. Lateral connectivity was \textit{hierarchical}, i.e. neuron i received lateral connections from neurons 1...i-1. This organization makes the network equivalent to Sanger's model, hence being capable of extracting principal components from data.
The model presented in \citep{kung1990} also used hierarchical lateral connectivity, with Hebbian/anti-Hebbian updates. Normalization was not performed explicitly, but it was achieved by means of Oja-like weight decay terms. Again, the resulting model was able to perform PCA.
Conversely, such an update scheme applied to networks with symmetric lateral connectivity, instead of hierarchical, as in \citep{leen1991}, provides a network that extracts the principal subspace. 
%In this work also the concept of \textit{weakly coupled} and \textit{full coupled} networks is introduced. A weakly coupled network has lateral/recurrent connectivity which modifies the network state in a single shot. In a full coupled network there are also lateral/recurrent connections, but the system is simulated as a continuous-time system which runs until convergence to a steady state is reached. The full coupled approach is more biologically plausible, in fact it was the one already in used in previous works, although the weak coupled counterpart (which can be considered as a single iteration approximation of the full coupled) is easier to run in software.
An interesting perspective on HaH is presented in \citep{seung2017}, which views Hebbian and anti-Hebbian learning parts as competing players in a game theoretic setting. 
Introducing nonlinearities in the learning process, with local learning rules in HaH networks, further enables the extraction of directions that maximize some generalized nonlinear moments of the data distribution \citep{karhunen1995}. 
Plasticity models such as BCM \citep{bienenstock1982} or successive variants \citep{intrator1992, law1994, brito2016}, have been shown to be effective to model receptive field formation in these scenarios.

A review of the above-mentioned approaches is provided in \citep{becker1996a}.
Although these methods provide biologically grounded mechanisms for subspace learning, their disadvantage is that, due to the recurrent nature of lateral connections, simulating these networks requires unfolding the recurrent dynamics in time. This requires a significant overhead compared to purely feedforward models such as Sanger's or Oja's. Nonetheless, we argue that the relationships between feedforward subspace learning models and HaH configurations have an important consequence, i.e. that the former are preferable for software simulation, while the correspondence with HaH models also provides biological support.
It is also possible to build neural networks capable of extracting minor components from data, i.e. eigenvectors associated with the smallest eigenvalues of the data covariance matrix. This is useful, for example, when we need to recover a signal buried in white noise. Minor component extraction can be achieved by reversing the sign of Oja's update rule, thus making it anti-Hebbian \citep{oja1992, luo1997}.
HaH networks have also been derived from learning objectives related to Classical Multi-Dimensional Scaling (CMDS - a.k.a. \textit{strain loss}, or \textit{similarity matching}) \citep{pehlevan2015a, pehlevan2015b, pehlevan2015c}, which are also shown to be related to subspace learning and principal component extraction. These approaches are also connected to manifold learning; therefore, they will be discussed in more detail in the next subsection.

\subsection{Manifold Learning Models}

Manifold learning models aim at mapping samples into lower dimensional spaces by constraining the output to preserve certain properties about the geometric structure of the data, beyond the simple linear relationships captured by methods such as PCA. Popular manifold learning methods include Isomap embeddings \citep{tenenbaum2000}, Locally Linear Embeddings (LLE) \citep{roweis2000}, or Laplacian eigenmaps \citep{belkin2003}.

An interesting manifold learning approach is represented by Classical Multi-Dimensional Scaling (CMDS) \citep{cox2008}. The reason is that recent work has derived HaH neural networks capable to optimize the CDMS objective \citep{pehlevan2015a, pehlevan2015b, pehlevan2015c}. The particular form of the objective, known as \textit{similarity matching}, or \textit{strain loss}, is the following:
\begin{equation} \label{eq:strain_loss}
    Y^* = \underset{Y}{arg \ min} \ \| X^T X - Y^T Y \|^2_F
\end{equation}
Where $X$ is a matrix obtained by concatenating a set of input vectors and similarly $Y$ is the matrix of the output vectors, while $\| \cdot \|_F$ is the Frobenius norm. For linear mapping, this objective is equivalent to standard subspace learning \citep{pehlevan2015a}. Let's give an intuitive interpretation of what this loss represents: $X^T X$ is a matrix whose elements are the dot products of pairs of input vectors, hence they represent the similarity of input vectors with other input vectors, and the same holds for $Y^T Y$. Therefore, that difference represents how much the similarity metric gets distorted when moving from the input space to the output space, and this is what should be minimized. The authors show that the problem can be solved by applying the following biologically-grounded neural dynamics and learning rules:
\begin{equation} \label{eq:strain_loss_hebb_updates} \begin{split}
& y = W \, x - M \, y \\
& \Delta W_{i, j} = \frac{y_i \, (x_j - W_{i, j} \, y_i)}{D_{i}} \\
& \Delta M_{i, j \neq i} = \frac{y_i \, (y_j - M_{i, j} \, y_i)}{D_{i}}, \ M_{i, i} = 0 \\
& \Delta D_i = y_i^2
\end{split} \end{equation}
where matrices $W$ and $M$ represent respectively the weights associated with the feed-forward and lateral interactions, while $D$ is a vector containing the cumulative squared activations of the neurons, which act in the equations as a dynamic learning rate.

As in the case of CMDS, manifold learning objectives appear to be effective principles for deriving data transformations that can be suitably mapped to neural layers, with potential applications for neuromorphic computation, thus representing an interesting open research area.

\subsection{Sparse Coding (SC)} \label{sec:plasticity:multineuron:sc}

Another interesting feature observed in biological networks is \textit{sparsity} in the neural activations, i.e. only a small percentage of neurons (around 1\%) activate simultaneously to encode a given stimulus \citep{lennie2003}. This property might derive simply from metabolic/energetic constraints, but it is also possible that sparsity plays a relevant role to support effective information encoding strategies \citep{foldiak1990, olshausen1996a}. Indeed, similar coding strategies are also observed in networks trained with backprop \citep{agrawal2014}.

The \textit{Sparse Coding} (SC) principle explicitly introduces a sparsity constraint in the learning framework. Let $\mathcal{X}$ be a dataset of input vectors. SC assumes that the elements of $\mathcal{X}$ can be represented as linear combinations of few basis vectors $\mathbf{d}_1, \mathbf{d}_2, ... \mathbf{d}_N$, also called \textit{words}. For a compact representation, let $D$ be a matrix whose columns are the word vectors, which is called the \textit{dictionary} matrix. The goal is to find, for each input $\mathbf{x} \in \mathcal{X}$, an encoding in terms of linear combinations of the dictionary vectors
\begin{equation}
    \mathbf{\hat{x}} = \mathbf{\hat{x}}(\mathbf{y}, D) = \sum_i y_i \mathbf{d}_i = D \, \mathbf{y}
\end{equation}
which minimizes the \textit{representation error}, i.e. a distance measure between the original and the reconstructed input. The vector of linear combination coefficients $\mathbf{y} = (y_1, y_2, ... y_N)^T$ is also called the \textit{code} vector. An additional constraint imposed by SC is the \textit{sparsity} of the representation $\mathbf{y}$. For example, using Euclidean distance as an error metric, the SC objective can be expressed as:
\begin{equation} \label{eq:sc_objective}
    \mathcal{L}_{SC}(\mathbf{y}, D) = \sum_{\frac{1}{2} \mathbf{x} \in \mathcal{X}} (\mathbf{x} - \mathbf{\hat{x}}(\mathbf{y}, D))^2 + \lambda \mathbf{C}(\mathbf{y}(\mathbf{x}))
\end{equation}
where $\mathbf{C}(\mathbf{y}) = (C(y_1), ..., C(y_N))^T$ is a cost function that penalizes dense codes, while $\lambda$ is a hyperparameter. In principle, we could choose function $C$ to simply count the number of non-zero elements of $y$, but this definition is not well suited for gradient-based optimization. Smother alternatives can be considered, such as $L_1$ or $L_2$ penalties on $\mathbf{y}$, but other forms are possible \citep{olshausen1996a, olshausen1996c}.

\begin{figure}
    \centering
    \begin{subfigure}{0.3\textwidth}
        \includegraphics[width=\textwidth]{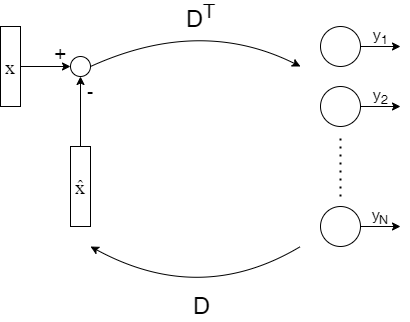}
        \caption{Sparse coding layer with error recirculation.}
        \label{fig:sc}
    \end{subfigure}
    ~
    \begin{subfigure}{0.3\textwidth}
        \includegraphics[width=\textwidth]{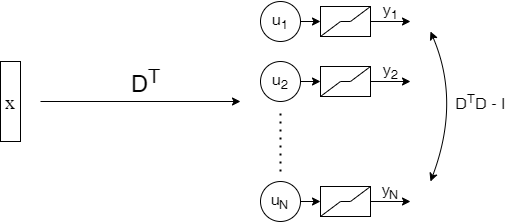}
        \caption{Sparse coding layer with feedforward and lateral connections, and a shrinkage-thresholding nonlinearity implementing the Locally Competitive Algorithm (LCA).}
        \label{fig:lca}
    \end{subfigure}
    \caption{Neural architectures for sparse coding layers.}
    \label{fig:sc_layers}
\end{figure}

\begin{figure}
    \centering
    \includegraphics[width=0.25\textwidth]{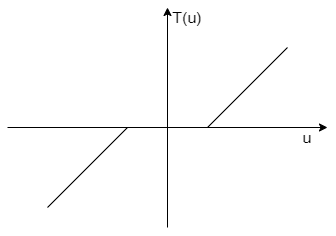}
    \caption{Shrink-thresholding nonlinearity.}
    \label{fig:shrink}
\end{figure}

Given the dictionary, there are various algorithms to find sparse codes, such as Orthogonal Matching Pursuit (OMP) \citep{pati1993}, or Fast Iterative Shrinkage-Thresholding Algorithm (FISTA) \citep{beck2009}. If the dictionary is not given a priori, other algorithms can be used to find the dictionary vectors, for example, based on clustering \citep{li2003, he2006a} or eigenvalue decomposition \citep{he2006b, aharon2006}. 
However, a neurally-grounded approach \citep{olshausen1996a} can be derived by considering SC as two nested optimization problems: first find optimal coding coefficients for a fixed dictionary, and then optimize the dictionary code vectors given the coding coefficients found before. In this scenario, the first optimization stage corresponds to unfolding the neural dynamics, while the second stage yields the synaptic dynamics. 
Specifically, starting from a code $\mathbf{y}$, we can minimize the objective in Eq. \ref{eq:sc_objective} by iterating gradient descent steps w.r.t. $\mathbf{y}$.This leads to the following update for $\mathbf{y}$: 
\begin{equation} \label{eq:sc_dynamics}
    \Delta \mathbf{y} \propto D^T \, (\mathbf{x} - \mathbf{\hat{x}}) - \mathbf{C}'(\mathbf{y})
\end{equation}
where $\mathbf{C}'(\mathbf{y}) = (C'(y_1), ..., C'(y_N))^T$ denotes the derivative of the sparsity-inducing cost function.
This formulation corresponds to a neural layer (Fig. \ref{fig:sc}) where the activations $\mathbf{y}$ represent the sparse code. The dictionary $D$ corresponds to feedback connections that map the code back to the reconstruction $\mathbf{\hat{x}} = D \, \mathbf{y}$, and the residual error $(\mathbf{x} - \mathbf{\hat{x}})$ is computed. The transpose dictionary $D^T$ represents forward connections that modify the activations $\mathbf{y}$ based on the previous error, such that the updates follow a gradient descent direction. The residual error recirculates for a number of iterations until convergence is reached, while $-C'(\mathbf{y})$ contributes to a sparsity-inducing decay in the activations. 
Once convergence is reached, the second phase of optimization involves the dictionary vectors. Calling $\mathbf{y}^*$ the sparse code after convergence, the dictionary can be optimized by another gradient descent step, w.r.t. $D$:
\begin{equation}
    \Delta D = \eta \, (\mathbf{x} - D \, \mathbf{y}^*) \, (\mathbf{y}^*)^T
\end{equation}
Note that Hebbian PCA algorithms can be considered as a special case of SC with a single iteration of the recirculation process (which is sufficient for convergence, under the assumption that the dictionary is orthogonal) and no sparsity constraint.

Although there is no explicit neural circuitry to support the error recirculation process, it is possible to show that sparse coding can also be implemented using once again feedforward and lateral connections, supporting biologically plausible local processing and plasticity (Fig. \ref{fig:lca}). 
By rewriting Eq. \ref{eq:sc_dynamics} as
\begin{equation}
    \Delta \mathbf{y} \propto D^T \, \mathbf{x} - D^T D \, \mathbf{y} - C'(\mathbf{y})
\end{equation}
we can observe two contributions in particular: $D^T \, \mathbf{x}$, which corresponds to a feedforward term, and $- D^T D \, \mathbf{y}$ which corresponds to a lateral interaction term, where $D^T$ and $- D^T D$ are respectively the feedforward and lateral connection weights, which replace the error recirculation process.
An SC neural layer can also be formulated in terms of an Energy-Based Model (EBM) \citep{hopfield1982, haykin}, as done in the Locally Competitive Algorithm (LCA) \citep{rozell2008}. An EMB is a dynamical system characterized by a state which evolves according to certain equations, in such a way that a certain \textit{energy} function is progressively reduced. In the SC case, neurons maintain an internal state $\mathbf{u}$, represented by their \textit{membrane potential}, in which stimuli are integrated over time. Outputs $\mathbf{y}$ are connected to $\mathbf{u}$ by a \textit{sparsifying} monotonic nonlinearity: $\mathbf{y} = T(\mathbf{u})$. 
The evolution of the system is described by the following equations:
\begin{equation}
    \begin{split}
        \Delta \mathbf{u} & \propto D^T \, \mathbf{x} - u - (D^T D - I) \, \mathbf{y} \\
        \mathbf{y} & = T(\mathbf{u})
    \end{split}
\end{equation}
which can be shown to minimize Eq. \ref{eq:sc_objective}, which is therefore the energy function of this EBM.
Also in this case, the system is characterized by a feedforward interaction $D^T \, \mathbf{x}$, and a lateral interaction $-(D^T D - I) \, \mathbf{y}$, where $D^T$ and $-(D^T D - I)$ are respectively the feedforward and lateral connection weights.
The specific form of the nonlinearity $T(\cdot)$ is related to the choice of the sparsity-inducing cost term $C(\mathbf{y})$ in Eq. \ref{eq:sc_objective} as follows \citep{rozell2008}:
\begin{equation}
    \lambda T^{-1}(y_i) = C'(y_i) + y_i \qquad i=1, ..., N
\end{equation}
A common choice is the shrinkage-threshold function (Fig. \ref{fig:shrink}), which corresponds to the $L_1$ sparsity penalty. This constitutes the LCA formulation of SC \citep{rozell2008}. As the name of the algorithm suggests, a form of \textit{local competition} take place among neurons. Indeed, it can be noticed that the nonlinearity, together with the lateral connections, induces a competitive interaction, where activations below a threshold are suppressed, while activations above a threshold inhibit the others.

\subsection{Independent Component Analysis (ICA)}

\begin{figure}
    \centering
    \includegraphics[width=0.5\textwidth]{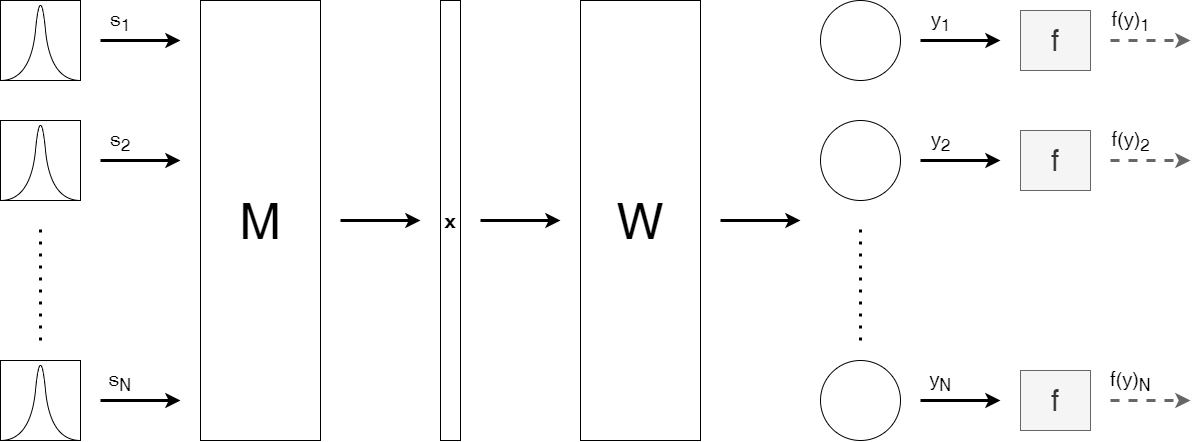}
    \caption{Blind Source Separation (BSS) problem. A mixing process $M$ generates samples $\mathbf{x}$ from source variables $s_1, ..., s_N$. A demixer $W$ mapping samples $\mathbf{x}$ to outputs $y_1, ..., y_N$. The goal is to find a demixer capable of reconstructing the original sources, without using information about the sources themselves.}
    \label{fig:bss}
\end{figure}

PCA looks for directions in the data space along have maximum variance while being maximally decorrelated. This idea can be generalized also to higher-order statistical moments. In particular, a stronger condition than decorrelation is represented by independence. Independent Component Analysis (ICA) \citep{hyvarinen} addresses the problem of finding data representations into a set of maximally independent variables. ICA has strong relationships with the Blind Source Separation (BSS) problem \citep{jutten1991}. In BSS, data are assumed to be generated by a mixing process as in Fig. \ref{fig:bss}. A \textit{mixer} generates data samples $\mathbf{x}$ from source variables $\mathbf{s} = (s_1, ..., s_N)^T$ (sampled from a problem-dependent distribution) through a \textit{mixing matrix} $M$: $\mathbf{x} = M \, \mathbf{s}$. A \textit{demixer} maps samples $\mathbf{x}$ to outputs $\mathbf{y} = (y_1, ..., y_N)^T$, through a \textit{demixing matrix} $W$: $\mathbf{y} = W \, \mathbf{x}$. The goal is to find a demixer that is capable of reconstructing the original sources using only information about the samples. The task is challenging because the information about the sources themselves is not available to guide the search for the desired demixer. 
Nonetheless, theoretical results show that the problem can be solved \citep{jutten1991, comon1991, comon1994, cardoso1996, cardoso1997, cardoso2001, cardoso2003, hyvarinen}, and the original sources can be correctly identified (up to a permutation), provided that the source distribution is non-Gaussian. 
Another scenario where the sources cannot be identified, without additional information, is represented by a nonlinear mixing process. Furthermore, depending on whether the number of sources is larger or smaller than the sample dimension (\textit{overcomplete} or \textit{undercomplete} mixtures, respectively), identifiability could be affected. Specifically, undercomplete mixtures are identifiable, while overcomplete mixtures require an additional constraint on the \textit{sparsity} of the sources. This condition is reminiscent of the SC problem, and indeed it has been shown that ICA and SC are actually related \citep{olshausen1996b}.

ICA approaches rely on information-theoretic methods to identify the correct demixer. A popular approach is the \textit{nautral gradient} method \citep{amari1996}, which enforces the independence requirement by minimizing the Mutual Information (MI) of demixer outputs. This can be achieved by considering the distribution of the demixer outputs, say $\mathbf{q}_\mathbf{Y}(\mathbf{y})$, and observing that the output variables are independent only if their distribution is a \textit{factorial} distribution, i.e. has the form $\mathbf{p}_\mathbf{Y}(\mathbf{y}) = \prod_{i=1}^N p_{Y_i}(y_i)$ (we will come back later on the choice of the specific form of $p$). Therefore, an objective can be defined as the minimization of the Kullback-Leibler (KL) divergence between the joint distribution $\mathbf{q_Y}$ and target factorial distribution $\mathbf{p_Y}$:
\begin{equation}
    \mathcal{L}_{ICA} = D_{KL}(\mathbf{q_Y}(\mathbf{y}) || \mathbf{p_Y}(\mathbf{y}))
\end{equation}
A gradient descent step on this objective w.r.t. the demixing matrix $W$ can be shown to lead to the following weight update equation:
\begin{equation}
    \delta W = \eta \, (W^{-T} - \mathbf{f}(\mathbf{y}) \, x^T)
\end{equation}
where $W^{-T}$ denotes the inverse transpose of $W$, and $\mathbf{f}(\mathbf{y}) = (f(y_1), ..., f(y_N))^T$ is a nonlinearity whose form is related to the choice of the target distribution $p$ as follows:
\begin{equation}
    f(y) = \frac{d}{dy} \log p(y) = \frac{p'(y)}{p(y)}
\end{equation}
The drawback of this learning rule is that it requires the computation of a matrix inversion $W^{-T}$ at each step, which is expensive. However, the method proposed in \citep{amari1996} suggests resorting instead to a natural gradient, which is a modification of the ordinary gradient by taking into consideration the underlying geometric structure of the problem. In our case, the natural gradient is simply obtained by multiplying the ordinary gradient by $W^T \, W$. Notice that the resulting direction will still be a descent direction on the objective because we are multiplying the gradient by a positive semi-definite matrix (so that the resulting vector will have a positive scalar product with/be less the 90° away from the gradient). Luckily, this multiplication removes the undesired inverse, leading to the following updated equation for ICA.
\begin{equation}
    \delta W = \eta \, (I - \mathbf{f}(\mathbf{y}) \, y^T) \, W
\end{equation}

\begin{figure}
    \centering
    \begin{subfigure}[t]{0.3\textwidth}
        \centering
        \begin{tikzpicture}
            \begin{axis}[
                axis x line=middle,
                axis y line=middle,
                xlabel = {$y$},
                ylabel = {$p_Y(y)$},
                ymax = 1.5,
                ymin = -0.5,
                width = \textwidth,
                ticks = none,
            ]
                \addplot[
                    domain=-5:5,
                    samples=60,
                ]{1.1*exp(-2*ln(cosh(x*1.2))) + 0.06};
                \addplot[
                    domain=-5:5,
                    samples=60,
                    dashed,
                ]{exp(-x*x*0.5) + 0.02};
            \end{axis}
        \end{tikzpicture}
        \caption{Super-Gaussian distribution (thick line) compared to Gaussian (dashed line).}
        \label{fig:ica_supergauss}
    \end{subfigure}
    ~
    \begin{subfigure}[t]{0.3\textwidth}
        \centering
        \begin{tikzpicture}
            \begin{axis}[
                axis x line=middle,
                axis y line=middle,
                xlabel = {$y$},
                ylabel = {$p_Y(y)$},
                ymax = 1.5,
                ymin = -0.5,
                width = \textwidth,
                ticks = none,
            ]
                \addplot[
                    domain=-5:5,
                    samples=60,
                ]{max(0.9*exp(-0.5*x*x*1.2*1.2+ln(cosh(x*1.2))) - 0.06, 0)};
                \addplot[
                    domain=-5:5,
                    samples=60,
                    dashed,
                ]{exp(-x*x*0.5) + 0.02};
            \end{axis}
        \end{tikzpicture}
        \caption{Sub-Gaussian distribution (thick line) compared to Gaussian (dashed line).}
        \label{fig:ica_subgauss}
    \end{subfigure}
    \\
    \begin{subfigure}[t]{0.3\textwidth}
        \centering
        \begin{tikzpicture}
            \begin{axis}[
                axis x line=middle,
                axis y line=middle,
                xlabel = {$y$},
                ylabel = {$f(y)$},
                ymax = 1.5,
                ymin = -1.5,
                width = \textwidth,
                ticks = none,
            ]
                \addplot[
                    domain=-3:3,
                    samples=35,
                ]{tanh(x)};
            \end{axis}
        \end{tikzpicture}
        \caption{Activation function for super-Gaussian distributions.}
        \label{fig:ica_tanh}
    \end{subfigure}
    ~
    \begin{subfigure}[t]{0.3\textwidth}
        \centering
        \begin{tikzpicture}
            \begin{axis}[
                axis x line=middle,
                axis y line=middle,
                xlabel = {$y$},
                ylabel = {$f(y)$},
                ymax = 1.5,
                ymin = -1.5,
                width = \textwidth,
                ticks = none,
            ]
                \addplot[
                    domain=-3:3,
                    samples=35,
                ]{x - tanh(x)};
            \end{axis}
        \end{tikzpicture}
        \caption{Activation function for sub-Gaussian distributions.}
        \label{fig:ica_tanh_compl}
    \end{subfigure}
    \caption{Types of distributions and corresponding nonlinearities in ICA.}
    \label{fig:ica_nonlinearities}
\end{figure}
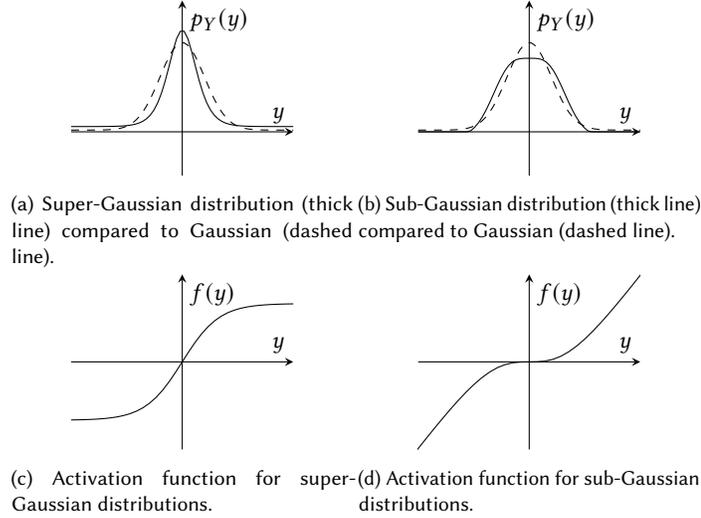

Let's go back to the choice of the distribution model for $p$. Ideally, the form of the target marginal distribution $p$ should correspond to the distribution of the sources, but this is often unknown in practice. However, domain expertise can help the expert to make assumptions about the distribution of the sources and choose $q$ accordingly. Distributions can be divided into two families: \textit{super-gaussian} and \textit{sub-gaussian}.
Super-gaussian distributions are characterized by a sharp central peak and heavier tails compared to a Gaussian. Examples of super-gaussian distributions are the \textit{Laplacian} distribution, whose log-density is $\log p_Y(y) = \alpha -|y|$ (where $\alpha$ is a normalizing constant), or the distribution with log-density $\log p_Y(y) = \alpha -2 \log \cosh (y)$. Sub-gaussian distributions are fat at the center and lighter-tailed compared to Gaussian, and some examples are the uniform distribution, or the distribution with log-density $\log p_Y(y) = \alpha - (\frac{1}{2} y^2 - \log \cosh (y))$.
Theoretical results \citep{hyvarinen1997a, hyvarinen1998b, hyvarinen, haykin} show that optimization will succeed as long as the assumed distribution belongs to the correct family as the true distribution. A simple condition to test whether a chosen nonlinearity is adequate for the source distribution of the given dataset is the following:
\begin{equation}
    \xi_i = \mathbb{E}[y_i \, f(y_i) - f'(y_i)] > 0 \qquad i = 1, ..., N
\end{equation}
Note that, if $\xi_i$ is negative for a given $f(y_i)$, then it will be positive for $g(y_i) = y_i - f(y_i)$  (assuming that data are normalized to zero mean and unit variance).
For instance, the example distributions $\log p_Y(y) = \alpha -2 \log \cosh (y)$ and $\log p_Y(y) = \alpha - (\frac{1}{2} y^2 - \log \cosh (y))$ lead to functions $f(y) = \tanh(y)$ and $f(y) = y - \tanh(y)$, respectively. A plot of super-Gaussian and sub-Gaussian distributions, and related nonlinearities, is shown in Fig. \ref{fig:ica_nonlinearities}.

The ICA approach is easily mapped into a feedforward neural layer (with architecture as Fig. \ref{fig:bss}), but alternative formulations also map into layers with feedforward and lateral connections \citep{jutten1991, fyfe1995, oja1996, karhunen1996, amari1998, hyvarinen1998b}. ICA is also related to nonlinear PCA methods \citep{karhunen1995, becker1996a}. The latter aims at canceling out some generalized higher-order cross-moments between the outputs (the specific form depends on the choice of the nonlinearity). Independence represents a stronger condition, as it implies the annulment of moments of all orders. However, nonlinear PCA approaches can also be effective in practice, as the minimization of higher-order moments often represents a good proxy to the maximization of independence.

Other approaches for ICA have been proposed in the literature, such as \textit{InfoMax} \citep{bell1995}, i.e. maximizing the mutual information between the demixer output filtered by a nonlinearity $f$ (Fig. \ref{fig:bss}), whose shape depends on the source distribution, and samples $\mathbf{x}$, or maximum-likelihood approaches \citep{pham1997}. Indeed, these various formulations can be shown to be equivalent \citep{pham1997, cardoso1997}. Another interesting aspect is the connection between ICA and SC \citep{olshausen1996b}. 
The SC problem can be mapped to a maximum-likelihood formulation by considering the SC objective function $\mathcal{L}_{SC}$ (Eq. \ref{eq:sc_objective}) as a log-likelihood of a probability density function:
\begin{equation}
    \ell_{\mathbf{Y}}(\mathbf{y}, D) = \frac{1}{Z} \, e^{-\mathcal{L}_{SC}(\mathbf{y}, D)} = \frac{1}{Z} \, e^{- \sum_{\frac{1}{2} \mathbf{x} \in \mathcal{X}} (\mathbf{x} - \mathbf{\hat{x}}(\mathbf{y}, D))^2 - \lambda \mathbf{C}(\mathbf{y}(\mathbf{x}))}
\end{equation}
where $Z$ is a normalizing constant. Without the sparsity-related term $\mathbf{C}$, this would just be a Gaussian density attributing higher probability density to lower reconstruction errors $(\mathbf{x} - \mathbf{\hat{x}})^2$. The sparsity term represents a prior that further penalizes dense codes by reducing the corresponding probability.
Indeed, the sparsity condition in SC is related to the form of the marginal distribution in ICA. In particular, super-gaussian distributions induce sparse representations, being characterized by samples that are either close to zero (often), or very large (rarely).

If samples come from a temporal process, another class of BSS approaches exists, which is able to identify sources based on the temporal structure of the signals \citep{matsuoka1995, belouchrani1997, meyer2001, choi2002a, choi2002b}. 
Finally, recent developments have provided a principled way to overcome previous theoretical limitations to the identifiability of nonlinear mixtures \citep{hyvarinen2016, hyvarinen2017, hyvarinen2019}. These approaches are able to achieve nonlinear ICA by augmenting the data with additional information, such as temporal information, or any other auxiliary variable available, and then training a model to distinguish between the \textit{true} augmented data and some \textit{negative} data with a randomized auxiliary variable. The idea of extracting information by contrasting positive and negative views of the data has also emerged in various other domains \citep{lai2001, sun2008, andrew2013, elmadany2016, dorfer2016, gatto2017, li2004, dorfer2015, koch2015, hoffer2015, ramachandran2017, baltruvsaitis2018, hossain2019, stefanini2022, kaur2021, oord2018, lowe2019, chen2020a}. This is the topic of \textit{multi-view} or \textit{contrastive learning}, which is discussed in the next subsection.

\subsection{Multi-View Learning Models}

\begin{figure}
    \centering
    \includegraphics[width=0.4\textwidth]{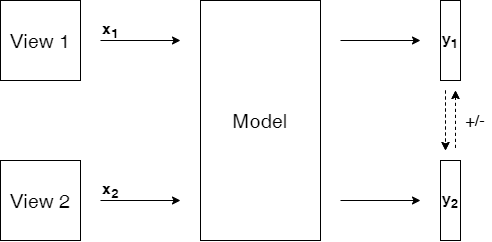}
    \caption{Schematic representation of a multi-view learning module. The module generates representations for a pair of inputs, driving positive pair together, and negative pairs far apart.}
    \label{fig:multiview}
\end{figure}

In the field of machine learning, a variety of approaches exist which are based on the idea of processing data observed from multiple \textit{views}. The concept of view is problem-dependent, but, for example, it can refer to multiple modalities through which the data are presented (visual, auditory, text, etc.), multiple copies of the same sample obtained by different transformations, multiple frames at different time instants, multiple features from a feature vector. Another example can be simply a sample and corresponding label information. The general idea is to map different related views (\textit{positives}) to a common representation while making sure that unrelated views (\textit{negatives}), which are typically obtained by a randomized pairing of views from different samples, are mapped to distinct representations (Fig. \ref{fig:multiview}).

Canonical Correlation Analysis (CCA) \citep{lai2001, sun2008, andrew2013, elmadany2016, dorfer2016, gatto2017}, max-margin approaches \citep{mao1993, pang2005, demir2005, li2004, dorfer2015, koch2015, hoffer2015}, multi-modal learning \citep{ramachandran2017, baltruvsaitis2018, hossain2019, stefanini2022, kaur2021}, contrastive representation learning \citep{oord2018, arora2019, henaff2019, lowe2019, chen2020a, bardes2021}, can all be considered as multi-view learning approaches.

CCA \citep{lai2001, sun2008, gatto2017} aims at finding linear projections of the different data modes so that the representations of related views are maximally correlated, while different views are uncorrelated. The \textit{IMax} approach \citep{becker1996b} is similar, but it considers the Mutual Information (MI) between different views instead of simple correlation. Discriminative CCA is a particular instantiation in which the views that are considered are a data view and a target to be predicted \citep{kim2007, sun2007, elmadany2016, dorfer2016}, which has been shown to be equivalent to linear regression methods \citep{shin2011}. Nonlinear extensions of CCA through deep mappings were also recently explored \citep{andrew2013, elmadany2016, dorfer2016}. Moreover, Hebbian neural network implementations of CCA methods also exist \citep{lai2001, gatto2017}.

Max-margin approaches \citep{mao1993, pang2005, demir2005, li2004, dorfer2015, koch2015, hoffer2015} are supervised methods where label information is used to derive a mapping of samples into a feature space, in order to optimize class separability. Linear Discriminant Analysis (LDA)  \citep{mao1993, pang2005, demir2005} is a classical method that minimizes the distance in feature space of same-class samples from the class centroid (\textit{intra-class} distance), and maximizes the distance between different class centroids (\textit{inter-class} distance). Equivalence between LDA and least-squares classification has been demonstrated \citep{ye2007}. 
While LDA focuses on linear mappings, max-margin methods were also extended to the nonlinear case \citep{santacruz1998, mika1999a, kim2005b, sugiyama2006, dorfer2015}.
A similar principle is also pursued in metric learning with siamese networks \citep{koch2015}, and triplet loss learning \citep{hoffer2015}, where the same class samples (positives) are mapped to nearby locations in feature space, while negative samples are far apart. 
Moreover, bio-inspired extensions of max-margin approaches for local learning have been proposed \citep{mao1993, demir2005, duan2021}.
It is worth mentioning that the idea of augmenting unsupervised methods with label information has been explored in discriminative clustering \citep{kaski2005, krause2010}, discriminative subspace learning \citep{bair2006, barshan2011, li2015, ritchie2019}, discriminative sparse coding \citep{mairal2008, mairal2009b, yang2011}, discriminative ICA \citep{akaho2002, bressan2002, dhir2011}, and discriminative manifold learning \citep{deridder2003, geng2005, zhang2008, wang2009b, raducanu2012, chien2016, liu2019a}.

Multi-modal learning \citep{ramachandran2017, baltruvsaitis2018, hossain2019, stefanini2022} aims at creating representations in DNNs that align different modalities of information, for example, images and text. Applications involve image captioning \citep{xu2015, sarto2023}, text-to-image synthesis \citep{reed2016, carrara2018a, zhang2017, zhang2018b}, and cross-modal retrieval \citep{messina2021b, wang2016b}. Hebbian learning could play a relevant role in reinforcing the observed correlations between different sensory pathways in the brain \citep{kaur2021}. Indeed, Hebbian approaches have been applied for cross-modal retrieval applications \citep{kaur2021}.

Contrastive representation learning \citep{oord2018, arora2019, henaff2019, lowe2019, chen2020a, bardes2021} is a popular approach for self-supervised learning in DNNs, which has shown promising results. Contrastive learning approaches aim at creating similar representations for elements that occur in similar contexts. This principle has been successfully used in the development of embedding models for language, such as the popular Word2Vec \citep{mikolov2013b}, where words that occur in similar contexts are mapped to similar vectors. Contrastive learning approaches took inspiration from this principle and popularized it also for images. In this case, any two neighboring image patches -- two views of the same input -- are required to have consistent representations, while unrelated patches should have distinct representations. The SimCLR \citep{chen2020a} approach uses instead differently augmented versions of the same images, imposing consistency conditions over the corresponding representations. When data also have a temporal dimension available, consistency between neighboring frames can also be an effective approach, and it has been shown to yield feature representations that resemble those of biological brains \citep{watanabe2018, zhuang2021}.

\section{Synaptic Plasticity Models in Deep Learning}  \label{sec:plasticity_dl}

\begin{figure}
    \centering
    \includegraphics[width=0.4\textwidth]{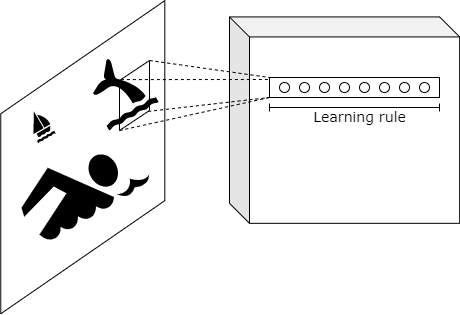}
    \caption{A given learning rule can be extended to the case of convolutional layers by applying it at different locations of the image, in a patch-wise fashion.}
    \label{fig:conv_learning}
\end{figure}

Synaptic plasticity learning rules can be extended to the case of convolutional layers by applying them at different locations of the images, in a patch-wise fashion, as illustrated in Fig. \ref{fig:conv_learning}.

The idea of developing feature extractors for CNNs, based on pattern discovery mechanisms such as SC or clustering, was already used in the work from Coates et al. \citep{stl10, coates2011a, coates2011b, coates2012}. An SVM classifier stacked on top of multiple convolutional layers (up to 3) was able to achieve 82\% accuracy on cifar-10 \citep{cifar}, 97\% on NORB \citep{norb}, 72\% con Caltech-101 \citep{caltech101}, 60\% on STL-10 \citep{stl10}. However, in order to achieve such results, these architectures require a significantly larger number of convolutional kernels (ranging from 1600 to 6000 in the proposed experiments) compared to traditional backprop-based architectures. For example, accuracy on CIFAR-10 drops to around 65\% with k-means-based learning and 100 convolutional filters \citep{stl10}.
In these works, it has been shown how 1st layer neurons tend to develop simple filters such as edges with various orientations. This also happens in networks trained with backprop and in biological brains. Nonetheless, the computational power of Hebbian neural networks trained with competitive schemes is reduced by the fact that different neurons tend to learn shifted versions of the same filter. This creates a correlation in the coding scheme which is not detected over the channel dimension but over the height and width dimensions of the feature map. In order to overcome this problem, the authors of \citep{dundar2015} introduced competition among neurons not only along the channel dimension but also with neighboring neurons in the height and width dimensions. The resulting 3-layer network achieved 74.1\% accuracy on the STL-10 \citep{stl10} dataset and 0.5\% error rate on MNIST.

In \citep{wadhwa2016a}, the authors propose a CNN architecture consisting of three convolutional layers, followed by an SVM classifier. The convolutional layers are trained, without supervision, to extract relevant features from the inputs. The proposed training algorithm, named \textit{Adaptive Hebbian Learning} (AHL), combines Hebbian weight update with k-WTA, pre-synaptic competition (given two winning neurons $j$ and $k$, a pre-synaptic neuron $i$ and the connecting synaptic weights $w_{i, j}$ and $w_{i, k}$, only the highest between $w_{i, j}$ and $w_{i, k}$ is updated) and dynamic recruiting/pruning of neurons. Additionally, a rule for learning bias terms, also used in previous works \citep{foldiak1989}, is adopted: the bias term is important to balance the activations of different neurons; hence, the idea is to keep a running average of neuron activations as $r$, choose a target activation value $A_{bias}$ and increase or decrease the bias in order to make $r$ approach $A_{bias}$. The rule, which is biologically motivated by homeostatic mechanisms \citep{turrigiano2012} for the stabilization of neural activity, is the following:
\begin{equation} \label{eq:rule_bias}
    \Delta b = \eta \, (r - A_{bias})
\end{equation}
The authors applied these ideas to different image datasets and obtained an error rate of 0.65\% on MNIST \citep{mnist}, an accuracy of 75.87\%  on CIFAR-10, and an error rate of 3.48\% on NORB.

Other works applied similar WTA-based approaches on fully connected networks. In \citep{krotov2019}, the authors achieved 98\% accuracy on MNIST and 50\% accuracy on CIFAR-10 with two fully-connected layers. In \citep{illing2019} various Hebbian models for sparse coding, principal, and independent component analysis were compared, in similar settings as \citep{krotov2019}. ICA models achieved the best results, with 53.9 \% accuracy on CIFAR-10, and 98.8\% mnist. PCA models achieved 50.8\% accuracy on cifar-10, and 98.2\% on MNIST, while SC achieved 50.2\% accuracy on cifar-10, and 98.4\% on MNIST.
The HaH model derived from the similarity matching criterion \citep{pehlevan2015a, pehlevan2015c} was applied to image classification tasks in \citep{bahroun2017}, again on the CIFAR-10 dataset, achieving 80\% accuracy with a single convolutional layer followed by an SVM classifier.
Hebbian learning approaches were further investigated in\citep{miconi2021}, where it was shown that deeper network layers were not able to develop more abstract features without supervision. In order to achieve more complex feature representations, this work proposed to introduce sparsity in the weight configuration by pruning some selected weights. Experiments with a 3-layer CNN show 64\% accuracy on CIFAR-10. This work also introduced a loss function formulation of the learning rules investigated, which allows a more immediate integration with modern deep learning frameworks.
The approaches in \citep{moraitis2021}, and layer \citep{journe2022}, use a soft-WTA training approach. In particular, the latter work shows for the first time increasing performance while going deeper with the number of layers, by resorting to very wide architectures where the number of neurons quadruples at each layer. It is shown that, as long as there are enough neurons, the network is able to disentangle the latent factors that describe the input, including those which provide the classification information. This work also contains experiments on ImageNet, but just for a single training epoch.
A very recent work \citep{gupta2022} provides an interesting investigation of Hebbian learning algorithms compared to backprop, showing superior performance of the former in single epoch training and in contexts of data scarcity.

The above-mentioned approaches only applied synaptic plasticity models to relatively shallow network architectures (generally with 2-3 layers). A further step was taken in \citep{lagani2019, mthesis, lagani2021d, lagani2022a, phdthesis}, where Hebbian WTA and PCA learning rules were investigated for training a 6-layer Convolutional Neural Network (CNN). Also, a supervised variant of Hebbian learning was proposed to train the final classification layer. Hybrid network models were also considered, in which some layers were trained using backprop and others using Hebbian learning. The results suggested that Hebbian learning is suitable for training early feature detectors, as well as higher network layers, but not very effective for training intermediate network layers. Furthermore, Hebbian learning was successfully used to retrain the higher layers of a pre-trained network, achieving results comparable to backprop, but requiring fewer training epochs, thus suggesting potential applications in the context of transfer learning (see also \citep{magotra2019, magotra2020, canto2020}).
Some contributions \citep{lagani2022b, lagani2022c} showed promising results of unsupervised Hebbian algorithms for semi-supervised network training, in learning scenarios with scarce data availability, achieving superior results compared to other backprop-based unsupervised methods for semi-supervised training such as Variational Auto-Encoders (VAE) \citep{kingma2013}. In further developments \citep{lagani2022b, lagani2022c}, a more efficient formulation of Hebbian learning was also proposed, that enabled the scaling up of experiments to complex image recognition datasets, such as ImageNet \citep{imagenet}, large-scale image retrieval, and complex network architectures, improving training speed up to a factor of 50. The solution, named \textit{FastHebb}, leveraged some observations that allowed us to rewrite Hebbian update equations in terms of matrix multiplications, to better exploit GPU acceleration.

\begin{table}
    \centering
    \caption{Experimental results of bio-inspired learning methods for deep learning applications on the CIFAR-10 dataset.}
    \label{tab:hebb_dl}
    \begin{tabular}{c|p{0.5\textwidth}|c}
         \textbf{Method} & \textbf{Description} & \textbf{CIFAR-10 Acc. (\%)} \\
         \hline
         K-means features & 3 conv layers with thousands of filters. & 82\%  \citep{coates2012}\\
         DHL & Competitive learning approach with 2 conv layers followed by SVM classifier, using label information to guide training. & 75.87\%  \citep{wadhwa2016b}\\
         Similarity matching & HaH network based on the similarity matching criterion \citep{pehlevan2015a}, with multi-scale filters, followed by an SVM classifier & 80\% \citep{bahroun2017} \\
         Krotov and Hopfield & Competitive learning approach with 2 fc layers. & 50\% \citep{krotov2019} \\
         Shallow PCA & Analysis of bio-inspired methods on shallow networks: PCA on a single fc layer + final classifier. & 50.80\%  \citep{illing2019} \\
         Shallow SC & Analysis of bio-inspired methods on shallow networks: PCA on a single fc layer + final classifier. & 50.20\%  \citep{illing2019} \\
         Shallow ICA & Analysis of bio-inspired methods on shallow networks: PCA on a single fc layer + final classifier. & 53.90\% \citep{illing2019} \\
         Hebbian learning with pruning & Objective function formulation of Hebbian approaches for gradient-based update computation. Hierarchical organization of Hebbian modules (3-layer CNN) with connection pruning and sparsification to induce more abstract features. & 64\% \citep{miconi2021} \\
         SoftHebb & Soft-WTA approach in deep CNNs (3 conv layers + final classifier). The number of filters is increased by a factor of 4 from each layer to the next, leading to very wide deep layers.  & 80.31\% \citep{journe2022} \\
         FastHebb& Semi-supervised Hebbian-backprop training based on Hebbian WTA/PCA. & 85\%  \citep{lagani2021b, lagani2021c, lagani2022c} \\
    \end{tabular}
\end{table}

Tab. \ref{tab:hebb_dl} summarizes the main experimental results presented above on the CIFAR-10 dataset.

We conclude this Section by mentioning some results related to the robustness properties of some bio-inspired models against adversarial perturbations \citep{szegedy2013, akhtar2018, yuan2019}. Early studies on adversarial attacks and defenses \citep{goodfellow2014b} already noticed that Radial-Basis Function (RBF) networks exhibited strong robustness against adversarial settings \citep{vidnerova2018, zadeh2018, goodfellow_cs231n_lecture}. For example, a Gaussian RBF activation function has the form:
\begin{equation}
    y(\mathbf{x}, \mathbf{w}) = e^{-\frac{|\mathbf{x}-\mathbf{w}|^2}{2 \sigma^2}}
\end{equation}
where the parameter $\sigma$ determines the width of the Gaussian, and the unit responds strongly only when the input $\mathbf{x}$ is within a certain distance from the reference weight vector $\mathbf{w}$. RBF-like activations could be biologically supported by frequency-dependent synaptic responses \citep{collingridge1988, markram1998}. Although RBF networks are hard to train, due to gradient vanishing problems \citep{goodfellow2014b}, gradient-free bio-inspired training methods could provide a useful mechanism to effectively leverage these types of models also in complex scenarios \citep{grossberg1976a, kohonen1982}. Lateral interaction and WTA-type nonlinearities have also proven useful to improve the adversarial robustness of DNN models \citep{kim2019a, xiao2019, panousis2021a, panousis2021b}.

\section{Spiking Neural Networks} \label{sec:spiking}

This Section introduces models of neural computation based on Spiking Neural Networks (SNNs) \citep{gerstner}, which more faithfully resemble real neurons compared to traditional ANN models. We start by introducing the various neuron models for SNN simulation. We highlight the applications related to biological and neuromorphic computing, which are of strong practical interest thanks to the energy efficiency of the underlying computing paradigm, and we discuss the challenges related to SNN training. We describe the biological plasticity models for spiking neurons and the connections with Hebbian synaptic plasticity.

\subsection{Spiking Neuron Models} \label{sec:spiking:models}

Spiking Neural Networks (SNNs) are a realistic model of biological networks \citep{gerstner, maass1997}.
While in traditional Artificial Neural Networks (ANNs), neurons communicate via real-valued signals, in SNNs they emit short pulses called \textit{spikes}. 
All the spikes are equal to each other and values are encoded in the timing or in the frequency with which spikes are emitted. 

Various spiking neuron models have been proposed in literature \citep{gerstner}: from the classical neuron description due to Hodgkin and Huxley (HH) \citep{hodgkin1952}, to more abstract but also computationally efficient models such as Izhikevich's \citep{izhikevich2003}, Spike-Response Models (SRM) \citep{gerstner1995}, and Leaky Integrate and Fire (LIF) \citep{abbott1993}. In particular, the LIF model is probably the highest-level description of spiking neurons, and also the most computationally efficient to simulate, which makes this model widely used in practice. 

LIF neurons behave like integrators, summing up all the received spikes (weighted by the synaptic coefficients) until a threshold is exceeded. At this point, an output spike is emitted. In practice, this integration logic is implemented in terms of an electric potential that is accumulated on the neural membrane every time an input spike is received; when the threshold is reached and the output spike is released, the neural membrane discharges the accumulated potential and the process restarts. Immediately after the discharge, the neuron enters its \textit{refractory period}, a time interval where it cannot spike regardless of its input. These units are leaky in the sense that, when no spikes are received in input, the membrane potential decays exponentially. 

\subsection{Neuromorphic Computing}  \label{sec:spiking:neuromorphic}

Thanks to the spike-based communication paradigm, biological neurons are extremely efficient in terms of energy requirements \citep{javed2010}. Energy efficiency is an important issue in modern deep learning \citep{badar2021}; hence, research is oriented toward different computing paradigms to support neural computation.

Spiking neuron models represent a promising computing paradigm due to the possible applications in the implementation of computing hardware that reproduces the behavior of biological neurons \textit{in silico}, with devices known as \textit{neuromorphic hardware} \citep{roy2019b, zhu2020, schuman2022, huynh2022, shrestha2022}. By reproducing the spike-based computation in hardware, researchers were able to develop extremely energy-efficient neuromorphic chips \citep{gamrat2015, wu2015}, such as Neurogrid \citep{benjamin2014}, TrueNorth \citep{merolla2014}, BrainScales \citep{schemmel2010, billaudelle2020}, Loihi \citep{davies2018}.
Despite the energy-efficient computing paradigm, SNN models have to face novel challenges, compared to traditional DNNs, related to the learning and optimization paradigms. In fact, traditional learning based on backprop is not adequate for SNN, because the spiking nonlinearity is not well suited for gradient-based optimization. Therefore, the following subsection is dedicated to the description of the counterpart model of Hebbian plasticity for SNNs: Spike Time Dependent Plasticity (STDP).

\subsection{Plasticity in SNNs}  \label{sec:spiking:plasticity}

In biological spiking neurons, learning occurs in the form of \textit{Spike Time Dependent Plasticity} (STDP) \citep{bi1998, song2000, gerstner}: when an input spike is received on a synapse and it is immediately followed by an output spike, then the weight on that synapse is increased. Specifically, a possible STDP rule can be expressed as follows:
\begin{equation} \label{eq:stdp}
\Delta w = \begin{cases}
    \eta^+ \, e^{-|\Delta t|/\tau^+} & \text{if $\Delta t > 0$} \\
    \eta^- \, e^{-|\Delta t|/\tau^-} & \text{otherwise}
\end{cases}
\end{equation}
where $w$ is the weight, $\Delta t$ is the time difference between the post-synaptic and the pre-synaptic spike, $\eta^+$ and $\eta^-$ are learning rate parameters ($\eta^+ > 0$ and $\eta^- < 0$) and $\tau^+$ and $\tau^-$ are time constants. Weight strengthening occurs when a pre-synaptic spike has been a likely cause for a post-synaptic spike, hence pre-synaptic and post-synaptic activations are correlated. If instead, a pre-synaptic spike occurred right after a post-synaptic one, then the two activations are anti-correlated and a weight decrease occurs. In this perspective, \textit{STDP realizes the Hebbian principle in the context of spiking neurons}. According to Eq. \ref{eq:stdp} The dependency between $\Delta w$ and $\Delta t$ according to the STDP weight update rule follows a double exponential profile.

\section{Concluding Remarks} \label{sec:concl}

We wish to provide a conclusion to this survey with some final remarks about the limitations and potentials of the bio-inspired methods presented so far. 

The main limitation of Hebbian and spiking models lies in the fact that their performance is not yet comparable to that of traditional DL approaches in terms of task performance. 
However, there exist several compelling factors that justify the study of biologically plausible learning models. Research on backprop-based models has witnessed extensive efforts, with the development of highly specialized hardware and solutions oriented to this type of optimization approach. Similarly, we anticipate a growth in efforts directed towards biologically plausible solutions in the near future, with an increasing interest in neuromorphic computing technologies  \citep{roy2019b, zhu2020, schuman2022, huynh2022, shrestha2022}. At the same time, additional efforts may lead to promising results both in the refinement of bio-inspired algorithms and in their application to more complex network architectures, and this work hopes to stimulate further interest in this sense.
Another disadvantage of biologically based models is related to the computational cost of simulating certain synaptic dynamics or unfolding the temporal evolution of complex neural circuitry. A recent work \citep{lagani2022c} addresses in part this problem by leveraging GPU parallelization more carefully.

A possible advantage of bio-inspired plasticity rules is \textit{locality}, in the sense that each layer of neurons can perform an update without having to wait for the whole network to process the input (each layer is independent of the subsequent ones). This is well suited for highly parallelizable layerwise training.
Additionally, local plasticity rules do not require gradient computations, which could make it easier to train deeper architectures without worrying about gradient vanishing problems; a possible application would be, for instance, efficient training of deep \textit{Radial Basis Function} (RBF) networks, which are of great interest for their adversarial robustness properties \citep{goodfellow2014b}, and whose computing model is biologically grounded \citep{collingridge1988, markram1998}. Training RBF networks with backpropagation is challenging because the gradient vanishes quickly when going far from the center of the RBF kernel. Therefore, an effective gradient-free alternative training approach would be helpful in these scenarios.

A key aspect of SNN and STDP models is the possibility of realizing energy-efficient neural network implementations in neuromorphic hardware, which could also find applications in embedded devices. Towards SNN training, where the backpropagation algorithm is not directly applicable, exploration of alternatives to backprop training, inspired for instance by biological plasticity mechanisms, represents a promising direction.

Finally, neuroscience and engineering fields can mutually influence each other, as neuroscience can provide engineers with valuable inspiration for the design of AI solutions, and, in turn,  technological advances can give insights to neuroscientists about what to look for in biological systems. Indeed, research efforts focused on the formulation of an algorithmic theory of the brain and further investigation of biologically plausible learning models are important to finally achieve a deeper understanding of how the human brain works, which could open possibilities of further advances both in technological and in medical fields \citep{lagani2021a, hbp}.

\begin{acks}
\noindent This work was partially supported by: \\
- Tuscany Health Ecosystem (THE) Project (CUP I53C22000780001), funded by the National Recovery and Resilience Plan (NRRP), within the NextGeneration Europe (NGEU) Program; \\
- Horizon Europe Research \& Innovation Programme under Grant agreement N. 101092612 (Social and hUman ceNtered XR - SUN project); \\
- AI4Media project, funded by the European Commission (H2020 - Contract n. 951911); \\
- INAROS (INtelligenza ARtificiale per il mOnitoraggio e Supporto agli anziani) project co-funded by Tuscany Region POR FSE CUP B53D21008060008.
\end{acks}

\bibliographystyle{ACM-Reference-Format}
\bibliography{references}

%\appendix

\end{document}